%% file: main.tex
\pdfoutput=1
\documentclass{article}

\PassOptionsToPackage{numbers, sort&compress}{natbib}
\PassOptionsToPackage{hyphens}{url}

\usepackage[eandd, final]{neurips_2026}

\usepackage[utf8]{inputenc} %
\usepackage[T1]{fontenc}    %
\usepackage{hyperref}       %
\usepackage{url}            %
\usepackage{booktabs}       %
\usepackage{amsfonts}       %
\usepackage{nicefrac}       %
\usepackage{microtype}      %
\usepackage{xcolor}         %

\input{macros}

\title{ViTeX-Bench: Benchmarking High-Fidelity Video Scene Text Editing}

\author{%
  Xinghao Chen$^{1}$ \quad Xiangbo Gao$^{1}$ \quad Jiongze Yu$^{1}$ \quad Yuheng Wu$^{1}$ \quad Zhengzhong Tu$^{1*}$ \\
  $^{1}$Texas A\&M University  \\
  $^{*}$\texttt{tzz@tamu.edu}  \\
  \url{https://vitex-bench.github.io/}
}

\begin{document}

\maketitle

\begin{abstract}
  Recent video generation is increasingly realistic and controllable, yet video editing remains comparatively underdeveloped, particularly for precise local edits that must preserve the original scene dynamics.
  Video scene text editing aims to replace text appearing on scene surfaces in a video, such as storefront signs, whiteboards, and product labels, while preserving the surrounding content, motion, and camera dynamics.
  Although scene text editing has been extensively studied for static images, video scene text editing that achieves high visual quality, temporal consistency, and edit locality remains largely underexplored. Existing resources offer limited paired real-video data, and general video-editing metrics do not directly measure whether the requested text remains correct over time. We introduce ViTeX-Bench, a benchmark suite comprising ViTeX-Dataset and a three-axis evaluation protocol. The dataset contains 387 real-world 720p videos with text-region masks and editing instructions: 230 provide reviewed, pipeline-generated paired edits for training, and 157 form a frozen evaluation split. The protocol evaluates text correctness, visual and temporal quality, and edit locality through 13 metrics, with one primary metric per axis and a Pareto comparison of their trade-offs. OCR calibration, human evaluation, and annotation-sensitivity analyses support the interpretation of these scores. Across eight baselines from four editing families, accurate text, temporal stability, and scene preservation remain difficult to achieve together. We also release ViTeX-Edit-14B, an open-source reference editor fine-tuned on the paired training split with motion-aligned glyph-video conditioning. It achieves CharAcc 0.714, the highest mean among the evaluated video-native editors, and with its glyph stream disabled it attains the lowest text-crop Warp among raw editor outputs. ViTeX-Bench provides a reproducible foundation for studying these trade-offs in video scene text editing.
\end{abstract}

\input{sec/1_intro}
\input{sec/2_related_work}
\input{sec/3_benchmark}
\input{sec/4_method}
\input{sec/5_experiment}
\input{sec/6_benchmarking_models}
\input{sec/7_limitations}
\input{sec/8_conclusion}

\begin{ack}
This work was supported in part by the GPU hardware provided to Texas A\&M University through the NVIDIA Academic Grant Program, in part by the Google Research Scholar Program, and in part by the Amazon Research Award.
\end{ack}

{
\small
\bibliographystyle{IEEEtran}
\bibliography{main}
}

\newpage
\appendix
\input{sec/X_suppl}

\end{document}

%% file: macros.tex
\usepackage{graphicx}
\usepackage{amsmath}
\usepackage{amssymb}
\usepackage{bm}
\usepackage{array}
\usepackage{tabularx}
\usepackage{multirow}
\usepackage{colortbl}
\usepackage{enumitem}
\usepackage{xspace}
\usepackage[noabbrev,nameinlink]{cleveref}

\crefname{section}{Sec.}{Secs.}
\Crefname{section}{Section}{Sections}
\crefname{table}{Tab.}{Tabs.}
\Crefname{table}{Table}{Tables}
\crefname{figure}{Fig.}{Figs.}
\Crefname{figure}{Figure}{Figures}
\crefname{appendix}{App.}{Apps.}
\Crefname{appendix}{Appendix}{Appendices}

\newcommand{\VX}{\texttt{ViTeX-\allowbreak{}Edit-\allowbreak{}14B}\xspace}
\newcommand{\VB}{\texttt{ViTeX-\allowbreak{}Bench}\xspace}
\newcommand{\VD}{\texttt{ViTeX-\allowbreak{}Dataset}\xspace}

\newlength\savewidth
\newcommand\shline{\noalign{\global\savewidth\arrayrulewidth \global\arrayrulewidth 1pt}\hline\noalign{\global\arrayrulewidth\savewidth}}

\definecolor{rankfirst}{HTML}{8FCFA1}
\definecolor{ranksecond}{HTML}{BFE3CB}
\definecolor{rankthird}{HTML}{E0F2E6}
\newcommand{\best}[1]{\cellcolor{rankfirst}#1}
\newcommand{\sbest}[1]{\cellcolor{ranksecond}#1}
\newcommand{\tbest}[1]{\cellcolor{rankthird}#1}

%% file: sec/1_intro.tex
\section{Introduction}
\label{sec:intro}

Recent video generation models~\cite{hunyuanvideo,cogvideox,ltxvideo,wan,vace,kling,moviegen,sora} have made substantial progress in producing photorealistic video clips with coherent motion, lighting, and geometry.
In practical editing workflows, however, users often need localized control rather than regenerating an entire video from scratch, or a combination of the two.
One common case is scene text editing, in which text is replaced on storefront signs, whiteboards, jerseys, screens, product labels, or other surfaces in the scene while leaving the rest of the video unchanged. This task is deceptively difficult. A successful edit must render the requested target string correctly, keep the edited text attached to the same surface as the camera or object moves, and preserve the surrounding appearance, lighting, and motion throughout the full clip.

\begin{figure*}[t]
  \centering
  \includegraphics[width=0.96\linewidth]{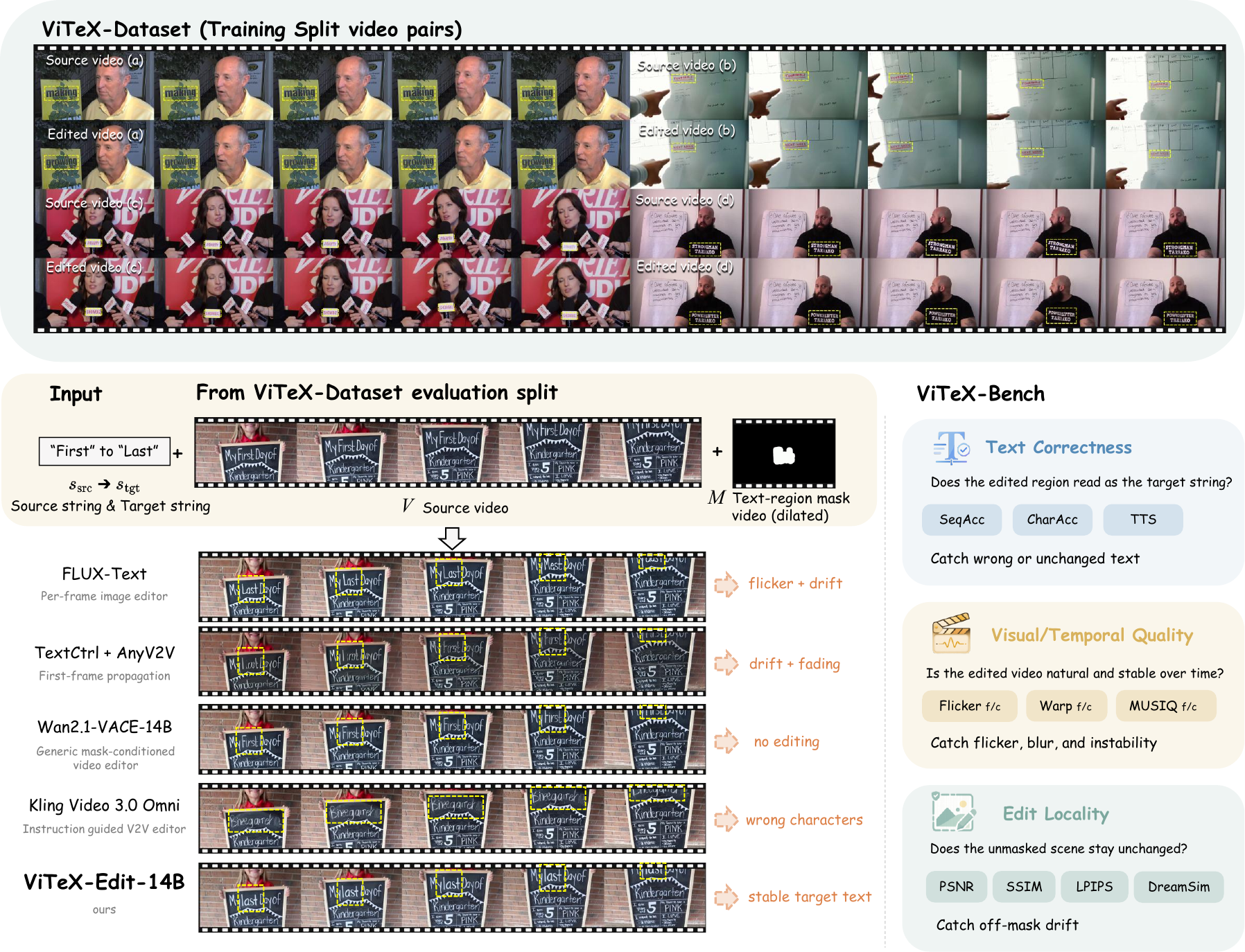}
  \caption{\textbf{Overview of ViTeX-Bench.} Paired training examples from ViTeX-Dataset (shown on top) illustrate the high visual fidelity of the paired data across diverse text-motion conditions. In ViTeX-Bench, each task instance provides a source video $V$, a text-region mask $M$, and a source-target string pair $(s_{\mathrm{src}}, s_{\mathrm{tgt}})$ (shown on the left). Representative baseline outputs (in the middle) exhibit distinct failure modes, while ViTeX-Bench (on the right) scores each edit along three axes (text correctness, visual quality, and edit locality) with a total of 13 metrics.}
  \label{fig:teaser}
\end{figure*}

Existing image and video editors~\cite{anytext2,textctrl,fluxtext,rsste,anyv2v,i2vgenxl,videopainter,vace} address different parts of this problem. Image scene-text editors such as FLUX-Text~\cite{fluxtext} can render accurate characters in individual frames, but independent edits introduce flicker and glyph drift. First-frame edit-and-propagate methods~\cite{anyv2v,i2vgenxl} extend a still-image edit through time, yet the inserted text can fade or drift over longer clips. Mask-conditioned and instruction-guided video editors, including VACE~\cite{vace}, VideoPainter~\cite{videopainter}, and Kling Video 3.0 Omni~\cite{kling}, model video dynamics but offer limited control over exact character sequences. Consequently, a stable video may retain the source text, a plausible edit may contain the wrong string, and individually correct frames may be temporally inconsistent (\cref{fig:teaser}).

Evaluating these failures requires task-specific data and measures. Related evidence from scientific chart editing shows that pixel similarity can miss semantic editing errors~\cite{figedit}. Image scene text editing has dedicated datasets and recognition metrics~\cite{srnet,anytext2,textctrl,rsste,glyphmastero,textdiffuser,anytext}, whereas paired edits of real-world videos remain limited. Existing video generation and editing benchmarks~\cite{vbench,vbench2,editboard,five,ivebench,vebench,vefx,li2026physics} assess instruction following, perceptual quality, temporal consistency, or preservation, but do not directly establish whether the edited region reads as the requested string throughout a video.

We introduce ViTeX-Bench to support systematic study of video scene text editing. ViTeX-Dataset contains 387 real-world 720p source videos from Panda-70M~\cite{panda70m} and InternVid~\cite{internvid}, with per-frame text-region masks and source--target string instructions. A human-in-the-loop pipeline produces paired edited references for 230 training videos; the remaining 157 form a frozen evaluation split.

Our primary contributions are the resource and its evaluation protocol. The dataset provides paired training examples and standardized evaluation inputs, with coverage statistics and annotation documentation. The protocol measures text correctness, visual and temporal quality, and edit locality through 13 complementary metrics. One primary metric per axis and a Pareto comparison make the trade-offs interpretable, while OCR calibration, human evaluation, and annotation-sensitivity analyses assess the reliability of the measurements.

To demonstrate the utility of the training split, we release ViTeX-Edit-14B as an open-source reference editor. It adapts a pretrained Wan2.1-VACE-14B backbone using a motion-aligned glyph-video stream that supplies target-character structure along the source text trajectory. Experiments with eight baselines across four editing families reveal distinct correctness, stability, and preservation failures. The reference editor combines the strongest mean character accuracy among the evaluated video-native editors with low temporal error, establishing a useful starting point for further work on this benchmark.

%% file: sec/2_related_work.tex
\section{Related Work}
\label{sec:related}

\paragraph{Video generation and editing.}
Video diffusion has evolved from pixel-space generation to large latent video models. Early pixel-space models extend image diffusion along a temporal axis~\cite{vdm,makeavideo,imagenvideo}. Latent-temporal models interleave temporal layers into pretrained image latent-diffusion backbones~\cite{videoldm,animatediff}. Native video diffusion transformers then learn spatio-temporal video distributions directly, ranging from open mid-scale systems~\cite{cogvideox,ltxvideo,opensora2} to multi-billion-parameter unified text--video models~\cite{hunyuanvideo,wan,moviegen,sora}. Editing methods built on top of these backbones largely follow two image-first patterns, namely per-video weight tuning~\cite{tunevideo} and training-free attention or feature propagation~\cite{fatezero,tokenflow,pix2video,controlvideo,videop2p,rave,slicedit,flowvid,ccedit,motiondirector,videoeditsurvey}. Image-to-video backbones~\cite{i2vgenxl,svd,videocrafter,dynamicrafter,lumiere} later supplied the propagation step for tuning-free first-frame editors~\cite{anyv2v}. Mask-conditioned and instruction-guided editors~\cite{videopainter,kling} extend these capabilities to localized and prompted edits. Sparse keyframe or reference conditioning has also been applied to instance insertion in PISCO~\cite{pisco}, whose released models our data pipeline builds on (\Cref{sec:pipeline}), as well as to video super-resolution~\cite{sparkvsr} and identity-preserving image-to-video generation~\cite{considgen}. ViTeX-Edit-14B focuses on character-level control, adding glyph-video conditioning for explicit character structure and temporal alignment.

\paragraph{Scene text editing.}
Image scene text editing began with GAN-era three-stage pipelines that disentangle background, foreground, and a learned text prior~\cite{srnet,swaptext,mostel}. Diffusion methods then introduced character-aware editors. GlyphDraw injects glyph priors through an image encoder~\cite{glyphdraw}, while DiffSTE and DiffUTE condition on dedicated character or OCR-based image encoders~\cite{diffste,diffute}. UDiffText, TextDiffuser, and TextDiffuser-2 unify these threads into character-aware diffusion frameworks and language-model-guided text painters~\cite{udifftext,textdiffuser,textdiffuser2}. More recent work extends this lineage with attribute-conditioned, structure--style-disentangled, recognition-supervised, FLUX-based, and OCR-free variants~\cite{anytext,anytext2,textctrl,rsste,fluxtext,textflux,textmastero}. GlyphMastero~\cite{glyphmastero} additionally shows that an explicit glyph encoder can supply stroke-level guidance to a diffusion editor, motivating the design of ViTeX-Edit-14B's conditioning pathway. On the video side, STRIVE~\cite{strive}, the closest predecessor, propagates a per-frame still-image edit on a small ROI-centric protocol. Concurrent text-to-video legibility work~\cite{vidtextpres} treats glyph quality as a generation-time concern, while LegiT~\cite{legit} evaluates text legibility in user-generated media rather than in-place editing. ViTeX-Bench focuses on in-place replacement in real videos, combining paired training data with frame-level recognition, temporal quality, and preservation metrics. Its reference editor adapts glyph-encoder conditioning to this temporal setting.

\paragraph{Benchmarks for video generation and editing.}
Video quality assessment aims to predict human judgments of perceived quality~\cite{vqasurvey}; COVER~\cite{cover}, for example, combines technical, aesthetic, and semantic quality estimates. Editing evaluation additionally requires checking the requested change and preservation of the source. General-purpose video benchmarks decompose quality into multiple primitive axes~\cite{vbench,vbench2,physiq}, but they target generation rather than instruction-guided editing. Several editing-specific suites have been introduced more recently. EditBoard~\cite{editboard}, FiVE~\cite{five}, and IVEBench~\cite{ivebench} adopt three-axis frameworks for instruction-guided edits. VE-Bench~\cite{vebench} pairs human MOS with a learned video-quality predictor, while OpenVE-3M~\cite{openve} provides million-scale instruction-conditioned editing data with three-aspect human ratings. TDVE-Assessor~\cite{tdvea} adapts large multimodal models as evaluators. VEFX-Bench~\cite{vefx} couples human annotations of instruction following, rendering quality, and edit exclusivity with VEFX-Reward, a learned evaluator conditioned on the source, instruction, and edited video. ViTeX-Bench complements these general editing evaluations with an OCR-anchored protocol for character-level correctness over time, coupled with temporal and locality measures on a frozen real-video split.

%% file: sec/3_benchmark.tex
\section{ViTeX-Bench: Dataset and Evaluation Suite}
\label{sec:bench}

\subsection{ViTeX-Dataset}
\label{sec:data}

\paragraph{Task formulation.}
Given a source video, a mask localizing the editable text region, and a source--target string pair, the task is to render the target string inside the mask while leaving the rest of the scene unchanged. We formalize each task instance as a tuple $(V, M, s_{\mathrm{src}}, s_{\mathrm{tgt}})$, where $V=\{f_t\}_{t=1}^{T}$ is a sequence of RGB frames $f_t\in\mathbb{R}^{H\times W\times 3}$, $M=\{m_t\}_{t=1}^{T}$ is a per-frame binary mask $m_t\in\{0,1\}^{H\times W}$ with $m_t=1$ on pixels inside the editable region, $s_{\mathrm{src}}$ is the character sequence visible inside that region, and $s_{\mathrm{tgt}}$ is the requested replacement. A method outputs an edited video $\hat V=\{\hat f_t\}_{t=1}^{T}$ satisfying these requirements. \VD instantiates this formulation at $T=120$ frames, $H\!\times\!W=720\!\times\!1280$, and $24$\,fps, releasing real-world source videos, masks, source--target string annotations, and paired edited videos on the training split.

\paragraph{Dataset overview.}
\VD{} contains $387$ real-world source videos manually screened from Panda-70M~\cite{panda70m} and InternVid~\cite{internvid}. The text therefore appears under natural lighting, surface geometry, and camera motion. The $230$-video training split provides $(V,\tilde V,M,s_{\mathrm{src}},s_{\mathrm{tgt}})$ tuples, where $\tilde V$ is the reviewed edit produced by our pipeline; the permanently frozen $157$-video evaluation split withholds $\tilde V$. Source and target strings are approximately length-matched within each pair and range from single characters to multi-word phrases across the dataset. The paired edits preserve the source context while providing target-text supervision (\Cref{fig:teaser}). They serve as training references rather than unique ground-truth renderings; their readability is examined in \Cref{app:human-validation}. Screening and composition statistics appear in \Cref{app:datasheet}.

\begin{figure}[t]
  \centering
  \includegraphics[width=\linewidth]{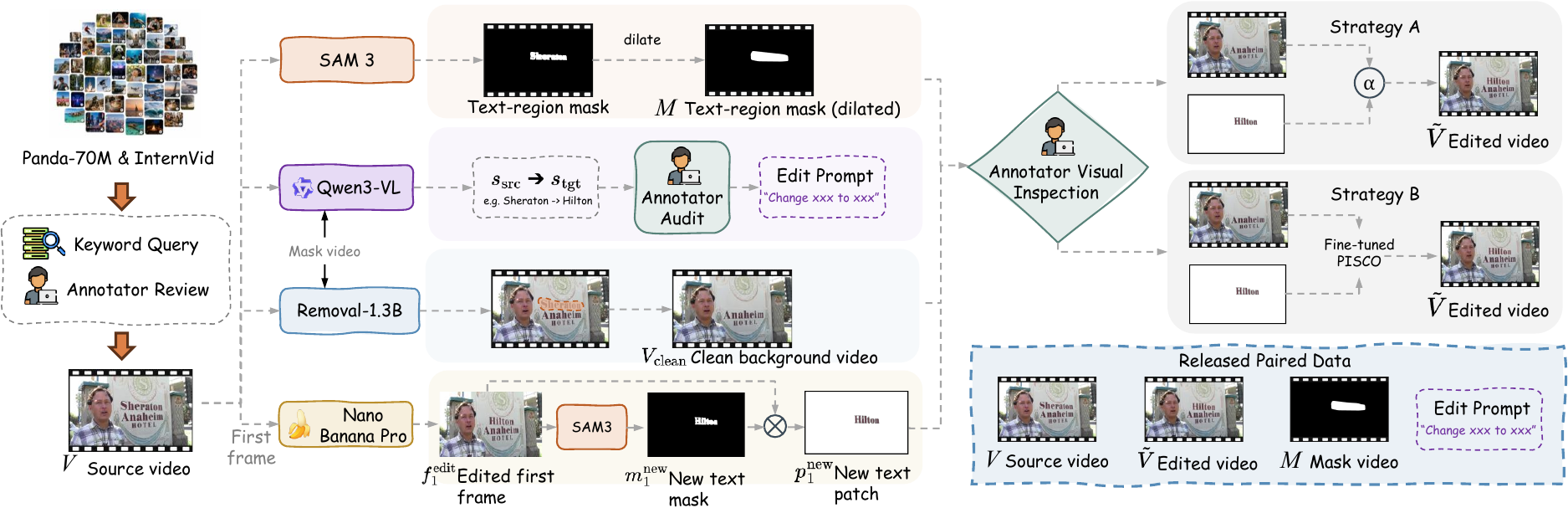}
  \caption{\textbf{Training data construction pipeline.} We compose the four assets ($M$, $(s_{\mathrm{src}},s_{\mathrm{tgt}})$, $V_{\mathrm{clean}}$, and $p_1^{\text{new}}$) into the paired edit $\tilde V$ via Strategy~A (alpha composition) or Strategy~B (PISCO-based inserter). The overview is in \Cref{sec:pipeline} and implementation details are in \Cref{app:pipeline}.}
  \label{fig:pipeline}
\end{figure}

\paragraph{Data construction pipeline.}
\label{sec:pipeline}
We draw source videos from Panda-70M and InternVid using keyword queries, then retain only videos suitable for editing and free of sensitive content. For each retained video, we construct four assets (\Cref{fig:pipeline}). The first is a dilated text-region mask $M$, annotated through a semi-automatic GUI built on the video segmentation model SAM~3~\cite{sam3}: an annotator marks the editable region with keypoints on the first frame, SAM~3 propagates the resulting mask to the remaining frames, and morphological dilation adds a margin around the glyph boundary. The second is a source--target string pair $(s_{\mathrm{src}},s_{\mathrm{tgt}})$, proposed by a vision-language model, Qwen3-VL-32B-Instruct~\cite{qwen3vl}, which reads $s_{\mathrm{src}}$ from the first-frame mask crop and suggests a similar-length $s_{\mathrm{tgt}}$; an annotator audits the result. The third is a clean background video $V_{\mathrm{clean}}$, produced by removal-1.3B~\cite{pisco}, a fine-tuned version of Wan2.1-VACE-1.3B that removes glyphs together with their cast shadows and highlights in the spirit of ROSE~\cite{rose}. The fourth is a first-frame target-text patch $p_1^{\text{new}}\!=\!f_1^{\text{edit}}\!\odot\!m_1^{\text{new}}$, where $f_1^{\text{edit}}$ is the first frame rewritten by an image editor, Gemini 3 Pro Image (Nano Banana Pro)~\cite{nanobananapro}, using the edit instruction from the earlier target-string generation step, and $m_1^{\text{new}}$ is the target-text mask from a second SAM~3 pass.

We adopted two strategies to account for the dynamics of scene text videos. \textbf{Strategy~A} alpha-composites $p_1^{\text{new}}$ onto each frame of $V_{\mathrm{clean}}$; it yields an edited video with minimal changes to the source but applies only when the text region remains static across all frames. \textbf{Strategy~B} uses a PISCO~\cite{pisco} inserter that takes $p_1^{\text{new}}$ as a first-frame reference and can therefore handle dynamic videos. We fine-tuned PISCO on an auxiliary scene-text insertion set with amodal-completion supervision. We visually classify each video as static or dynamic: dynamic videos use Strategy~B exclusively, while static videos run both strategies and retain the higher-quality output. The final paired training split contains $56$ Strategy-A videos and $174$ Strategy-B videos. Full pipeline details appear in \Cref{app:pipeline}.

\paragraph{Coverage and annotation reliability.}
\Cref{tab:coverage} summarizes the dataset's script, length, and typography coverage. The evaluation split covers four scripts: Latin, Chinese, Japanese, and Cyrillic. The coverage audit reports a mask-area ratio of $0.032\pm0.022$ and approximately $25\%$ static versus $75\%$ dynamic videos, based on visual motion classification. These motion categories differ from the training pipeline's $56/174$ strategy counts because static videos can use either construction strategy.

\begin{table}[t]
\centering
\caption{\textbf{\VD{} statistics.} All clips are $1280\!\times\!720$, 120 frames at 24\,fps. String lengths are in characters (mean\,$\pm$\,std); source and target strings range over 1--41 and 1--36 characters. Scripts counts the writing systems in the frozen evaluation split (Latin, Chinese, Japanese, Cyrillic). Font styles are shares of all 387 videos, rounded.}
\label{tab:coverage}
\small
\begin{tabular*}{\linewidth}{@{\extracolsep{\fill}}cccccccc@{}}
\toprule
\multicolumn{3}{c}{Videos} & \multicolumn{2}{c}{String length} & \multicolumn{3}{c}{Font style (\%)} \\
\cmidrule(lr){1-3} \cmidrule(lr){4-5} \cmidrule(lr){6-8}
Train & Eval & Scripts & Source & Target & Printed & Handwritten & Artistic \\
\midrule
230 & 157 & 4 & $8.0\pm5.7$ & $8.0\pm5.4$ & 23 & 44 & 33 \\
\bottomrule
\end{tabular*}
\end{table}

One author-annotator performed the original construction. An independent annotator repeated the mask pipeline on 12 difficulty-stratified clips, obtaining mask IoU $0.95$, Dice $0.98$, and crop-box IoU $0.94$. Under the alternative masks, DreamSim-loc and text-crop Warp rankings have Kendall $\tau=0.94$ and $1.00$, respectively. \Cref{app:coverage-validation} details this pilot audit and its scope.

\subsection{ViTeX-Bench Evaluation Suite}
\label{sec:bench-proto}

\paragraph{Evaluation protocol.}
\label{sec:metrics}
\VB{} evaluates outputs on the frozen $157$-video split along three axes: text correctness, visual and temporal quality, and edit locality. Its 13 core metrics probe these axes at complementary spatial scopes and perceptual sensitivities. We report one primary metric per axis together with the complete diagnostic vector (\Cref{sec:primary,sec:main}). Supplementary background-motion and identity probes extend the analysis without changing the core protocol.

\paragraph{Text correctness.}
We run an OCR recognizer, PP-OCRv5~\cite{ppocrv5}, on the dilated-mask crop of each source frame $f_t$ and predicted frame $\hat f_t$, producing normalized strings $s_t$ and $\hat s_t$, respectively. OCR backend configuration, confidence thresholding, and string normalization are detailed in \Cref{app:metrics-impl}. Source text is not always readable: motion, occlusion, or blur can obscure it on individual frames. We score correctness on \emph{source-detectable} frames to reduce confounding by source unreadability, and calibrate residual recognition errors in \Cref{app:human-validation}. To compare two strings, we use substring edit distance $d_{\mathrm{sub}}(r,c)$, defined as the minimum number of character edits required to transform reference $r$ into any contiguous substring of candidate $c$. Unlike standard Levenshtein distance, unmatched prefixes and suffixes of $c$ are free; a correct target embedded inside a longer OCR string is therefore not penalized for surrounding characters. \Cref{app:metrics-impl} gives a worked example. The induced similarity is $\mathrm{Sim}(r,c)=1-d_{\mathrm{sub}}(r,c)/\max(|r|,1)\in[0,1]$. The source-detectable set contains every frame on which the source-frame OCR string matches $s_{\mathrm{src}}$ at least halfway:
\begin{equation*}
\mathcal{D}=\{t:\mathrm{Sim}(s_{\mathrm{src}},s_t)\ge 0.5\},\qquad
\mathcal{P}=\{(t,t+1):t,t+1\in\mathcal{D}\},
\end{equation*}
and $\mathcal{P}$ collects the consecutive pairs inside $\mathcal{D}$ for temporal consistency. The three text-correctness primitives are
\begin{equation}
\begin{aligned}
\mathrm{SeqAcc} &= \operatorname*{mean}_{t\in\mathcal{D}} \mathbf{1}[d_{\mathrm{sub}}(s_{\mathrm{tgt}},\hat s_t)=0], \\
\mathrm{CharAcc} &= \operatorname*{mean}_{t\in\mathcal{D}} \mathrm{Sim}(s_{\mathrm{tgt}},\hat s_t), \\
\mathrm{TTS} &= \operatorname*{mean}_{(t,t+1)\in\mathcal{P}} \mathbf{1}[\hat s_t = \hat s_{t+1}],
\end{aligned}
\label{eq:correctness}
\end{equation}
where $\mathbf{1}[\cdot]$ is the indicator function. SeqAcc demands an exact substring match to $s_{\mathrm{tgt}}$; CharAcc gives partial credit for near-correct renderings; and TTS (temporal text stability) measures whether the decoded string is stable across adjacent detectable frames. TTS intentionally measures stability rather than correctness, so it must be read together with SeqAcc and CharAcc. Edge cases are handled in \Cref{app:metrics-impl}.

\paragraph{Visual quality.}
We score visual quality at two spatial scopes: the full output frame ($S=\mathrm{full}$) and a text-crop region ($S=\mathrm{crop}$). The crop is one static bounding box per video---the axis-aligned box enclosing the spatial union of every per-frame mask $\bigcup_{t=1}^{T} m_t$, enlarged by a fixed margin. For videos whose text moves across the scene, the box widens to cover the trajectory; every frame is still cropped through the same window, so $\mathrm{Flicker}_c$ and $\mathrm{Warp}_c$ measure glyph drift rather than bounding-box jitter (\Cref{app:metrics-impl}). Let $x_t^S$ denote the pixels of $\hat f_t$ inside scope $S$, and let $\mathcal{T}_S$ be the frame index set over which MUSIQ is averaged. We use $\mathcal{T}_{\mathrm{full}}=\{1,\dots,T\}$ for full-frame MUSIQ and $\mathcal{T}_{\mathrm{crop}}=\mathcal{D}$ for crop MUSIQ, so crop quality is averaged only when the source text is detectable. The six visual primitives are
\begin{equation}
\begin{aligned}
\mathrm{Flicker}_S &= \operatorname*{mean}_{t=1}^{T-1} \mathrm{MAE}(x_{t+1}^S, x_t^S), \\
\mathrm{Warp}_S    &= \operatorname*{mean}_{t=1}^{T-1} \mathrm{MAE}\!\left(x_t^S,\, \mathcal{W}(F^{\mathrm{src}}_{t\to t+1}, x_{t+1}^S)\right), \\
\mathrm{MUSIQ}_S   &= \operatorname*{mean}_{t\in\mathcal{T}_S} \mathrm{MUSIQ}(x_t^S),
\end{aligned}
\label{eq:visual}
\end{equation}
where $F^{\mathrm{src}}_{t\to t+1}$ is RAFT~\cite{raft} forward flow on the source video, $\mathcal{W}(F,x)$ backward-warps $x$ to frame $t$ using $F$, and MUSIQ~\cite{musiq} estimates perceptual quality without a reference. Flicker measures raw adjacent-frame differences; Warp compensates for source motion. Both can decrease under smoothing or nearly constant outputs, so their interpretation also requires correctness and perceptual quality. Full-frame variants capture global artifacts, while text-crop variants emphasize the edited region.

\paragraph{Edit locality.}
Edit locality measures how well a method preserves pixels outside the editable region. We construct a locality-only prediction $\hat f_t^{\mathrm{loc}}$ that retains predicted pixels outside the mask and substitutes source pixels inside, then average a per-frame metric $\mu$ over the video:
\begin{equation}
\begin{aligned}
\hat f_t^{\mathrm{loc}} &= (1-m_t)\odot \hat f_t + m_t\odot f_t, \\
\mu_{\mathrm{loc}}        &= \operatorname*{mean}_{t=1}^{T} \mu(\hat f_t^{\mathrm{loc}}, f_t), \quad \mu \in \{\mathrm{PSNR}, \mathrm{SSIM}, \mathrm{LPIPS}, \mathrm{DreamSim}\}.
\end{aligned}
\label{eq:locality}
\end{equation}
Inside the mask, $\hat f_t^{\mathrm{loc}}=f_t$, so differences arise from the unedited region. PSNR and SSIM~\cite{ssim} are higher-is-better similarities; LPIPS~\cite{lpips} and DreamSim~\cite{dreamsim} are lower-is-better perceptual distances. Pixel-level measures respond to small VAE reconstruction differences, whereas learned distances capture perceptual changes. We denote locality DreamSim by $\mathrm{DreamSim}_{\mathrm{loc}}$ (DreamSim-out). These framewise measures are complemented by background-motion and identity probes in \Cref{app:background-validation}. Implementation and edge cases appear in \Cref{app:metrics-impl}.

\paragraph{Primary metrics and comparison.}
\label{sec:primary}
The primary metrics are SeqAcc ($\uparrow$), $\mathrm{Warp}_c$ ($\downarrow$), and $\mathrm{DreamSim}_{\mathrm{loc}}$ ($\downarrow$), representing correctness, temporal quality, and locality. We compare their trade-offs through the Pareto set: a method is dominated when another is at least as good on all three and strictly better on one. The remaining ten metrics provide diagnostic detail. We report raw outputs separately from Composite post-processing and omit VideoPainter from temporal comparisons because of its adaptation pipeline (\Cref{sec:baselines}). Rankings summarize mean scores; confidence intervals quantify their uncertainty. We use no weighted aggregate across the three axes.

%% file: sec/4_method.tex
\section{ViTeX-Edit-14B}
\label{sec:method}

\VX{} adapts a pretrained video editor to the paired training split through motion-aligned character conditioning. It provides an open reference for the benchmark while retaining the backbone's pretrained video prior.

\begin{figure}[t]
  \centering
  \begin{minipage}[c]{0.5\linewidth}
    \centering
    \includegraphics[width=\linewidth]{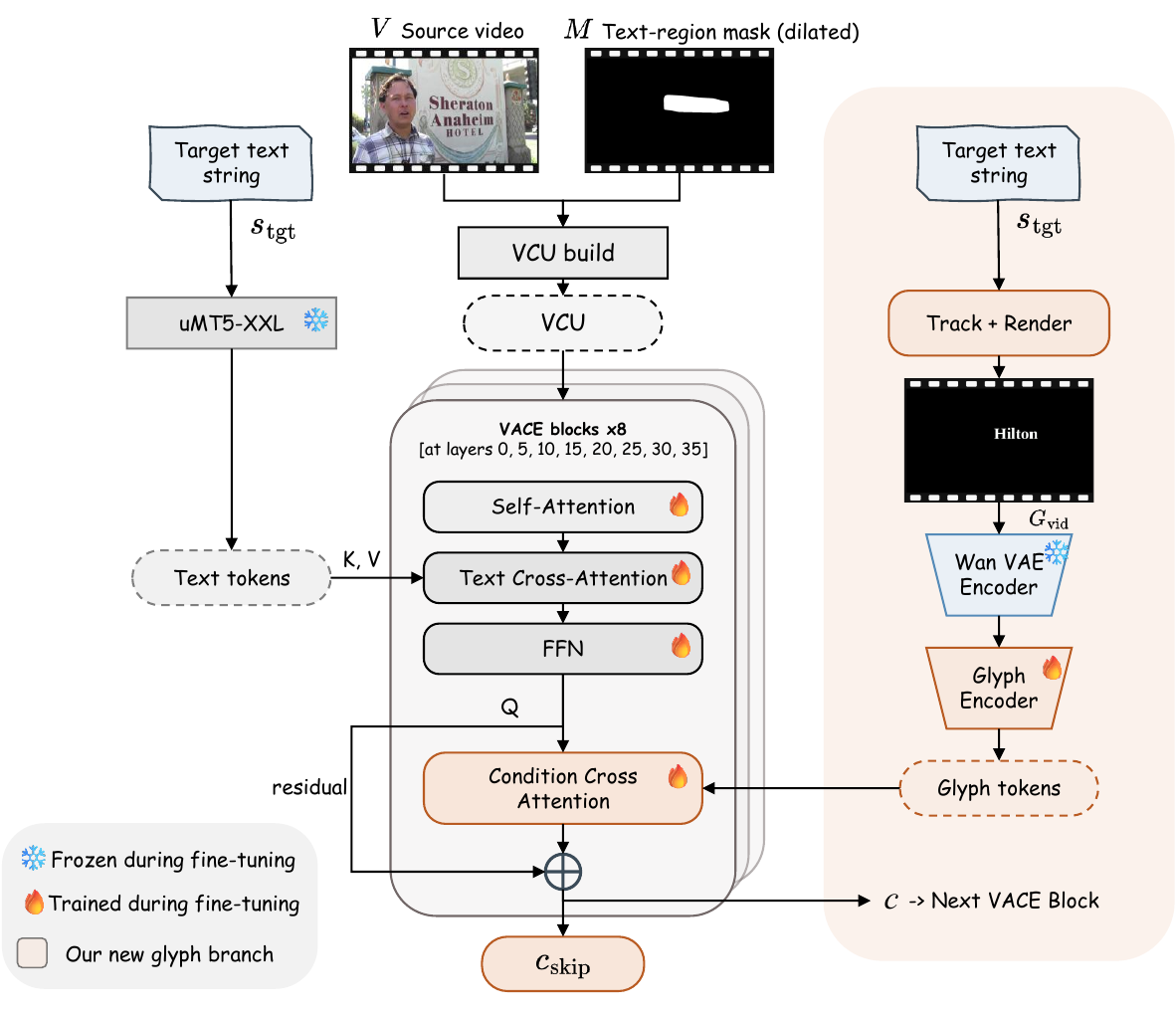}
  \end{minipage}\hfill
  \begin{minipage}[c]{0.46\linewidth}
    \caption{\raggedright \VX architecture. Three streams condition the VACE backbone: target text $s_{\mathrm{tgt}}$ via frozen uMT5-XXL, source $V$ and mask $M$ via the VCU, and a target-text glyph video pooled by the glyph encoder into tokens $E_G$. Every VACE block queries $E_G$ through an added condition cross-attention layer. Implementation details are in \Cref{app:vitex14b-impl}.}
    \label{fig:arch}
  \end{minipage}
\end{figure}

Wan2.1-VACE-14B~\cite{vace} provides two conditioning streams: a text encoder for $s_{\mathrm{tgt}}$ and a Video Condition Unit (VCU) for the source video and mask. We add a third stream: a target-text glyph video $G_{\mathrm{vid}}$ that supplies both character structure and source-aligned motion. To build $G_{\mathrm{vid}}$, we render $s_{\mathrm{tgt}}$ as a white-on-black glyph image in a typeface chosen to match the source font, detect the source-text quadrilateral inside the mask, track it across the remaining frames with a mask-based fallback, and projectively warp the glyph image with the resulting per-frame homographies, so $G_{\mathrm{vid}}$ follows the source text's position, scale, and perspective (typeface selection, OCR detector, and tracker in \Cref{app:vitex14b-impl}). We first tried two simpler alternatives: using $G_{\mathrm{vid}}$ at inference without fine-tuning, and routing it through the existing VCU during fine-tuning. Both yielded poor character correctness in qualitative pilots, motivating a dedicated glyph branch. \Cref{sec:glyph-ablation} ablates the glyph stream at inference; training-time component ablations remain future work.

\paragraph{Architecture.}
We build \VX{} on Wan2.1-VACE-14B and inherit its VCU together with the frozen uMT5-XXL text encoder (\Cref{fig:arch}). For the new glyph-video stream, a frozen Wan VAE encodes $G_{\mathrm{vid}}$ to a latent $z_G$. A stride-$(1,2,2)$ patch embedding flattens $z_G$ into tokens, and $64$ learnable queries $Q_{64}$ perform cross-attention pooling to produce a fixed-length glyph token bundle:
\begin{equation}
E_G = W_\mathrm{out}\cdot\mathrm{CrossAttn}\big(Q_{64},\,\mathrm{LayerNorm}(z_G)\big),
\label{eq:glyph-pool}
\end{equation}
where $W_\mathrm{out}$ is a zero-initialized output projection and the LayerNorm is a pre-norm applied to the keys and values before attention. Every VACE block then queries $E_G$: its hidden state $h$ passes through a lightweight condition cross-attention layer added back via a zero-initialized residual,
\begin{equation}
h' = h + W_o\cdot\mathrm{FlashAttn}\big(W_q\,\mathrm{LayerNorm}(h),\,W_k E_G,\,W_v E_G\big),
\label{eq:cond-xattn}
\end{equation}
where $W_q$, $W_k$, and $W_v$ are the query, key, and value projections and $W_o$ is a zero-initialized output projection. The zero-initialized residual projection preserves the backbone output at initialization. During fine-tuning we freeze the main DiT trunk, uMT5-XXL, and Wan VAE, updating only the VACE branch, the glyph encoder, and the condition cross-attention layers. Training follows a two-stage Flow-Matching supervised fine-tuning (SFT) curriculum ($576$ GPU-hours on $8\!\times\!\text{H100~80GB}$), and inference runs a single $50$-step pass. Per-stage hyperparameters and module dimensions appear in \Cref{app:vitex14b-impl}.

\paragraph{Shared Composite post-processing.}
Composite is a deterministic, training-free post-processing wrapper applicable to any editor. It matches the predicted region to the source through annulus-based LAB color transfer, then blends the region onto the source with a 4-pixel feathered boundary. This separates text-region synthesis from background reconstruction. We apply the same wrapper to all eight baselines and \VX{}, reporting its effects separately from raw model performance. The algorithm and re-scoring scope are given in \Cref{app:composite}.

%% file: sec/5_experiment.tex
\section{Experiments}
\label{sec:exp}

\subsection{Experimental Setup}

\paragraph{Baselines.}
\label{sec:baselines}
We compare eight baselines from four editing families, adapting their outputs to the common $1280\!\times\!720$, $120$-frame, $24$ fps evaluation grid. Full configurations are in \Cref{app:baselines}.

\textbf{Family A: per-frame image editing.} AnyText2, TextCtrl, FLUX-Text, and RS-STE~\cite{anytext2,textctrl,fluxtext,rsste} edit each frame independently; the outputs are concatenated into a video.

\textbf{Family B: first-frame editing and propagation.} TextCtrl edits the first frame, and AnyV2V~\cite{anyv2v} propagates it using the I2VGen-XL backbone~\cite{i2vgenxl}.

\textbf{Family C: mask-conditioned video inpainting.} Wan2.1-VACE-14B~\cite{vace} and VideoPainter~\cite{videopainter} receive the text-region mask and a prompt specifying the target string. VideoPainter's CogVideoX 1.0 backbone~\cite{cogvideox} requires spatial resizing and linear-blend temporal upsampling. The latter alters adjacent-frame residuals, so its Flicker and Warp scores are marked $\dagger$ and excluded from temporal rankings.

\textbf{Family D: instruction-guided video editing.} Kling Video 3.0 Omni~\cite{kling} receives the source video and a fixed editing-instruction template.

Family B represents first-frame propagation within the broader literature on tuning-free diffusion-based video editing~\cite{fatezero,tokenflow,pix2video,controlvideo,videop2p,rave,slicedit,flowvid,ccedit,motiondirector}. Evaluating additional systems requires method-specific adaptation to the target-string and long-video protocol (\Cref{app:related}).

\subsection{Main Quantitative Results}
\label{sec:main}

\Cref{tab:main} reports video-level mean scores for the eight baselines and the reference editor with and without its glyph stream. Text correctness uses source-detectable frames; five clips with no such frames are excluded from SeqAcc and CharAcc. Rankings compare these means, with 95\% video-bootstrap confidence intervals in \Cref{tab:text-ci,tab:vq-ci,tab:loc-ci}.

\begin{table}[!ht]
\centering
\caption{\textbf{Main evaluation results on \VB.} The 95\% bootstrap CIs are deferred to \Cref{tab:text-ci,tab:vq-ci,tab:loc-ci}. Per-column shading among raw editors only: \best{best}/\sbest{2nd}/\tbest{3rd} distinct displayed values; rounded ties share shading. \VX{} (text-only) runs the same weights with the glyph stream disabled at inference, so the target string comes only from the text prompt (\Cref{sec:glyph-ablation}). Composite rows are unranked post-processing controls; all-baseline controls appear in \Cref{tab:composite-all}. $f{/}c$ denotes full-frame/text-crop. $^{\dagger}$VideoPainter $\mathrm{Flicker}_{f{/}c}$ and $\mathrm{Warp}_{f{/}c}$ excluded from ranking (\Cref{app:baselines}). The Source video row ($\hat V = V$) is an unranked reference.}
\label{tab:main}
\footnotesize
\setlength{\tabcolsep}{2pt}
\renewcommand{\arraystretch}{0.95}
\resizebox{\linewidth}{!}{
\begin{tabular}{lc|ccc|cccccc|cccc}
\shline
& & \multicolumn{3}{c|}{\textbf{Text correctness}} & \multicolumn{6}{c|}{\textbf{Visual quality}} & \multicolumn{4}{c}{\textbf{Edit locality}} \\
\cmidrule(lr){3-5} \cmidrule(lr){6-11} \cmidrule(lr){12-15}
\textbf{Method} & \textbf{Fam.}
& SeqAcc$\uparrow$ & CharAcc$\uparrow$ & TTS$\uparrow$
& Flicker$_f\downarrow$ & Flicker$_c\downarrow$ & Warp$_f\downarrow$ & Warp$_c\downarrow$ & MUSIQ$_f\uparrow$ & MUSIQ$_c\uparrow$
& PSNR$\uparrow$ & SSIM$\uparrow$ & LPIPS$\downarrow$ & DreamSim$\downarrow$ \\
\hline
Source video                          & --- & 0.000 & 0.317 & 0.760 & 3.72 & 3.68 & 1.46 & 1.27 & 70.33 & 45.12 & $\infty$ & 1.000 & 0.000 & 0.000 \\
\hline
AnyText2~\cite{anytext2}              & A & 0.280 & 0.633 & 0.382 & \sbest{3.34} & 4.95 & 2.04 & 3.95 & 66.68 & 41.65 & 25.56 & 0.905 & 0.091 & 0.043 \\
TextCtrl~\cite{textctrl}              & A & \sbest{0.475} & \sbest{0.734} & 0.511 & 3.80 & 4.29 & \tbest{1.59} & 2.09 & \tbest{70.32} & 42.77 & \best{41.14} & \best{0.994} & \best{0.008} & \best{0.004} \\
FLUX-Text~\cite{fluxtext}             & A & \best{0.528} & \best{0.737} & 0.326 & 5.11 & 14.81 & 3.03 & 13.01 & 70.26 & 43.85 & 31.49 & 0.975 & 0.029 & \tbest{0.012} \\
RS-STE~\cite{rsste}                   & A & 0.354 & 0.626 & 0.534 & 3.73 & \tbest{3.66} & 1.61 & 1.81 & 69.57 & 34.26 & \sbest{37.00} & \sbest{0.983} & \tbest{0.024} & \sbest{0.007} \\
\hline
TextCtrl + AnyV2V~\cite{textctrl,anyv2v} & B & 0.057 & 0.308 & 0.257 & 4.98 & 4.98 & 4.11 & 3.97 & 69.41 & 33.85 & 21.08 & 0.785 & 0.225 & 0.073 \\
\hline
Wan2.1-VACE-14B~\cite{vace}           & C & 0.000 & 0.298 & \best{0.689} & 3.78 & 3.84 & 1.69 & \sbest{1.56} & \sbest{70.54} & \tbest{45.26} & \tbest{35.21} & \tbest{0.976} & \sbest{0.022} & \sbest{0.007} \\
VideoPainter$^{\dagger}$~\cite{videopainter} & C & \tbest{0.364} & 0.619 & 0.606 & 2.38\rlap{${}^{\dagger}$} & 2.62\rlap{${}^{\dagger}$} & 2.93\rlap{${}^{\dagger}$} & 3.35\rlap{${}^{\dagger}$} & 67.16 & 40.59 & 28.56 & 0.915 & 0.104 & 0.024 \\
\hline
Kling Video 3.0 Omni~\cite{kling}     & D & 0.000 & 0.208 & 0.641 & 4.25 & 4.08 & 3.12 & 2.90 & \best{72.23} & \best{47.75} & 21.18 & 0.843 & 0.176 & 0.061 \\
\hline
\textbf{\VX}                     & --- & 0.363 & \tbest{0.714} & \sbest{0.658} & \tbest{3.35} & \sbest{3.58} & \sbest{1.58} & \tbest{1.62} & 70.29 & \sbest{46.22} & 29.70 & 0.953 & 0.060 & 0.020 \\
\textbf{\VX\ (text-only)}         & --- & 0.339 & 0.688 & \tbest{0.653} & \best{3.29} & \best{3.45} & \best{1.56} & \best{1.55} & 69.81 & 43.84 & 29.25 & 0.953 & 0.058 & 0.023 \\
\hline
\textbf{\VX\ (Composite)}             & --- & 0.361 & 0.718 & 0.664 & 3.75 & 3.92 & 1.52 & 1.61 & 70.39 & 46.84 & 42.67 & 0.992 & 0.007 & 0.003 \\
\textbf{\VX\ (text-only, Composite)} & --- & 0.347 & 0.691 & 0.654 & 3.74 & 3.83 & 1.51 & 1.56 & 70.37 & 45.09 & 42.67 & 0.992 & 0.007 & 0.003 \\
\shline
\end{tabular}}
\end{table}

The three axes reveal distinct strengths: per-frame editors achieve the highest character accuracy, the reference editor attains the lowest temporal error, and bounding-box-local methods preserve the surrounding scene particularly well.

\paragraph{Text correctness.}
FLUX-Text and TextCtrl lead text correctness, reaching SeqAcc $0.528$ / $0.475$ and CharAcc $0.737$ / $0.734$. Wan2.1-VACE-14B and Kling both score SeqAcc $0$, reflecting unchanged or incorrectly rendered text. Among video-native editors, \VX{} achieves the highest mean CharAcc at $0.714$, compared with VideoPainter's $0.619$ ($+0.095$, or $15.3\%$ relative). VideoPainter has slightly higher SeqAcc ($0.364$ vs.\ $0.363$), showing that improved partial character accuracy does not necessarily yield more exact strings. Their intervals overlap; these differences describe observed means rather than established pairwise significance. TTS supplies a separate stability signal: the Source row scores $0.760$ despite never making the requested edit.

\paragraph{Visual quality.}
In its text-only configuration, \VX{} has the lowest mean Flicker$_f$, Flicker$_c$, Warp$_f$, and Warp$_c$ among comparable raw outputs; the glyph stream raises them slightly in exchange for higher text correctness. FLUX-Text combines high correctness with large text-region residuals (Flicker$_c=14.81$, Warp$_c=13.01$), consistent with its independent per-frame edits. Kling leads both MUSIQ measures despite SeqAcc $0$, illustrating the distinction between visual polish and successful text replacement. VideoPainter's temporal scores remain unranked because of its interpolation-based adaptation.

\paragraph{Edit locality and the Composite control.}
Bounding-box-local editors copy most exterior pixels from the source before encoding, whereas full-frame editors reconstruct the surrounding scene. Composite isolates this difference: for \VX{}, it raises PSNR-loc from $29.70$ to $42.67$ dB and reduces DreamSim-loc from $0.020$ to $0.003$, while SeqAcc changes from $0.363$ to $0.361$. Applied to all eight baselines, the same wrapper brings PSNR-loc to approximately $43$ dB and full-frame Flicker toward the source value $3.72$ (\Cref{tab:composite-all}). The improvement across methods indicates that source-pixel restoration accounts for much of the locality gain.

\paragraph{Primary-metric trade-offs.}
FLUX-Text, TextCtrl, RS-STE, both \VX{} configurations, and Wan2.1-VACE-14B form the Pareto set on the three primaries (\Cref{tab:pareto}). Wan2.1-VACE-14B remains non-dominated at SeqAcc $0$, demonstrating that membership alone does not establish editing success. The full diagnostic vector is therefore needed to interpret each operating point.

\paragraph{Glyph-stream ablation.}
\label{sec:glyph-ablation}
Disabling the glyph stream at inference (the text-only configuration), with all other weights unchanged, lowers SeqAcc from $0.363$ to $0.339$ and CharAcc from $0.714$ to $0.688$, while Warp$_c$ decreases from $1.62$ to $1.55$ (\Cref{tab:glyph-ablation}). A checkpoint-loading bug made this the configuration behind the submitted \VX{} results (\Cref{app:correction}).

\subsection{Calibration and Robustness Analyses}
\label{sec:validation}

\textbf{OCR and human evaluation.} On detectable source frames, OCR achieves exact-match accuracy $0.851$ and CharAcc $0.966$, with TTS $0.760$. Method-blinded human transcription agrees with the OCR-based method ranking at Spearman $\rho=0.95$. Three non-author raters also evaluated 70 video outputs on 1--3 scales. Their ordinal Krippendorff agreement is $0.87$ for text, $0.80$ for temporal quality, and $0.37$ for locality. Mean ratings correlate with SeqAcc, text-crop Warp, and DreamSim-loc at $+0.71$, $-0.40$, and $-0.53$, respectively ($p<0.001$). Text-crop Warp aligns more closely with temporal ratings than full-frame Warp ($-0.40$ vs.\ $-0.20$), supporting its selection as the temporal primary. Study protocols and limitations are detailed in \Cref{app:human-validation}.

\textbf{Scale and coverage.} Bootstrap resampling of the 152 source-detectable clips ($1{,}000$ replicates, seed $2064$) yields mean Kendall $\tau=0.925$ against the full SeqAcc ranking and retains the leading method in $95\%$ of size-152 replicates. In the non-Latin slice detailed in \Cref{app:coverage-validation}, AnyText2 leads with SeqAcc $0.168$ and CharAcc $0.295$; all other editors have CharAcc below $0.19$. Difficulty stratification and the independent mask audit appear in \Cref{app:coverage-validation}.

\textbf{Background preservation.} Supplementary BG-Warp, DINOv2-drift, and ArcFace probes of background motion, feature stability, and face identity broadly support the locality trends while revealing differences that the core metrics can obscure (\Cref{app:background-validation}).

%% file: sec/6_benchmarking_models.tex
\section{Diagnostic Failure Analysis}
\label{sec:breakdown}

\begin{figure}[t]
  \centering
  \includegraphics[width=0.96\linewidth]{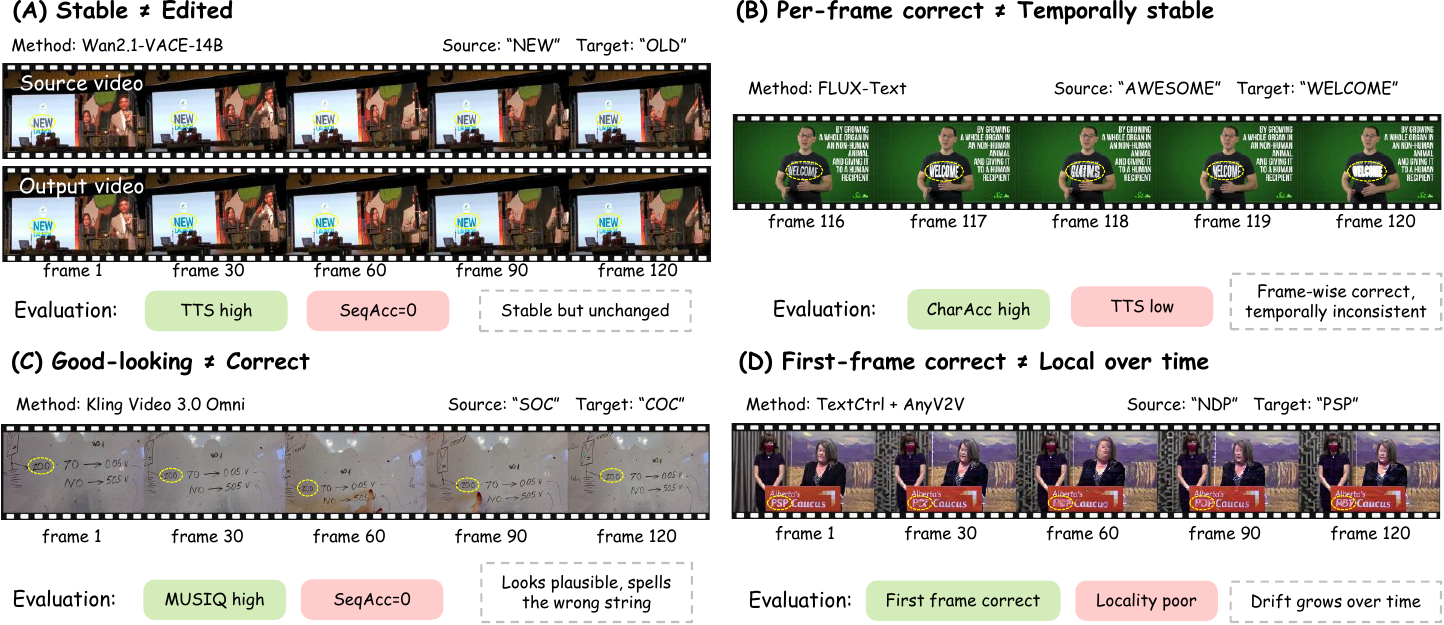}
  \caption{Four representative failure cases. \textbf{(A)} Wan2.1-VACE-14B: masked region returned essentially as the source, target string never rendered. \textbf{(B)} FLUX-Text~\cite{fluxtext}: per-frame editing yields legible text on individual frames, but the glyph identity is inconsistent between adjacent frames. \textbf{(C)} Kling Video~3.0~Omni~\cite{kling}: high per-frame visual quality, but the rendered text does not match the requested target. \textbf{(D)} TextCtrl{+}AnyV2V~\cite{textctrl,anyv2v}: the first-frame edit propagates while the rendered text and surrounding scene structure drift away.}
  \label{fig:qual}
\end{figure}

\Cref{fig:qual} connects four qualitative failures to their metric signatures: Wan2.1-VACE-14B retains the source text (high TTS, zero SeqAcc), FLUX-Text is legible but temporally unstable, Kling is polished yet incorrect, and TextCtrl+AnyV2V drifts in both text and background. Correctness, temporal quality, and locality must therefore be inspected jointly.

\begin{figure}[t]
  \centering
  \includegraphics[width=0.96\linewidth]{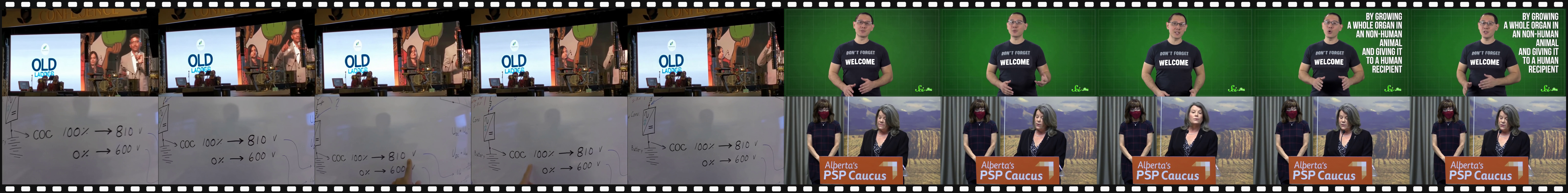}
  \caption{\VX{} outputs for the four source videos in \Cref{fig:qual}, shown with five evenly spaced frames per video. Panels (A)--(D) correspond to the examples used to illustrate Wan2.1-VACE-14B, FLUX-Text, Kling Video~3.0~Omni, and TextCtrl{+}AnyV2V, respectively. On these selected examples, \VX{} renders the target text correctly and maintains it across the displayed frames.}
  \label{fig:vx-qual}
\end{figure}

On the same videos, \VX{} renders the target strings and maintains them across the displayed frames (\Cref{fig:vx-qual}); \Cref{tab:main} quantifies performance over the full split.

%% file: sec/7_limitations.tex
\section{Limitations}
\label{sec:limitations}

\textbf{Coverage and scale.} ViTeX-Dataset focuses on localized, readable, predominantly Latin-script text. Handwritten and artistic styles are represented, but dense layouts, curved surfaces, severe occlusion, extreme motion, and non-Latin scripts remain sparsely covered. Bootstrap stability characterizes the sampled domain, while broader coverage requires additional data. The reference editor demonstrates useful adaptation from 230 paired videos; beyond the inference-time glyph ablation, training-time component ablations and training-scale studies remain future work.

\textbf{Measurement and annotation.} OCR accuracy depends on script, style, and readability, so source calibration provides context rather than a universal target-text ceiling. Five evaluation clips fall outside SeqAcc/CharAcc support. Human calibration comprises a transcription study conducted by one author and a three-rater study of 70 outputs, with limited locality agreement ($\alpha=0.37$). Independent mask annotation covers 12 clips and uses the same propagation pipeline; other annotation stages lack independent agreement studies. The background and face probes extend preservation analysis but do not cover arbitrary object identity or semantics.

\textbf{Construction and provenance.} Paired edits are reviewed outputs of a model-assisted pipeline and may retain rendering errors. Per-record target-string rejection/resampling counts and first-frame editing retry rates were not logged in the initial release. Broader independent audits, richer provenance, and expanded linguistic and geometric coverage would strengthen future versions.

%% file: sec/8_conclusion.tex
\section{Conclusion}
\label{sec:conclusion}

ViTeX-Bench provides paired training data and a frozen evaluation protocol for video scene text editing. Its three-axis design makes character correctness, temporal quality, and scene preservation explicit, while calibration and robustness analyses clarify how to interpret the measurements. Across eight baselines and the ViTeX-Edit-14B reference editor, the results expose distinct failure modes and persistent trade-offs. Shared Composite controls further separate synthesis quality from background restoration. Together, the dataset, evaluation code, and reference editor provide a reproducible basis for measuring progress in video scene text editing. Release URLs are listed in \Cref{app:release}.

%% file: sec/X_suppl.tex
\section{Released Resources}
\label{app:release}

The five resources below are publicly released: a GitHub Pages project site, the dataset and model weights on the Hugging Face namespace \url{https://huggingface.co/ViTeX-Bench}, the code for the benchmark and for \VX{} at \url{https://github.com/taco-group/ViTeX-Bench}, and a GitHub Pages leaderboard.

\begin{itemize}[leftmargin=*,itemsep=2pt]
\item \textbf{Project page.} \url{https://vitex-bench.github.io/}. A single landing page organizes side-by-side qualitative comparisons (the source video with a translucent text-region mask alongside outputs from every baseline and \VX{} / \VX\,(Composite)), a static mirror of the leaderboard, the per-metric table, the architecture and dataset-construction figures, and the four representative baseline failures from \Cref{sec:breakdown}. The four components below are linked from the page header for direct navigation.

\item \textbf{\VD{}} (dataset, $387$ source videos; $230$ paired training videos; CC-BY-NC 4.0). \url{https://huggingface.co/datasets/ViTeX-Bench/ViTeX-Dataset}. The $230$-video training split is distributed as full $(V,\tilde V,M,s_{\mathrm{src}},s_{\mathrm{tgt}})$ tuples; the $157$-video evaluation split is permanently frozen and withholds $\tilde V$. Datasheet, Croissant 1.0 metadata, and dataset license are co-located on the dataset repository (\Cref{app:datasheet}).

\item \textbf{\VB{} evaluation code} (Apache-2.0). \url{https://github.com/taco-group/ViTeX-Bench}. This repository contains implementations of the 13 metrics with the frozen recognizers and metric definitions, plus a single-command runner that downloads the evaluation split on first run.

\item \textbf{\VX{}} (Apache-2.0). Weights: \url{https://huggingface.co/ViTeX-Bench/ViTeX-Edit-14B}. The open-source reference editor fine-tuned on the $230$-video training split. Its code lives in the \mbox{\texttt{vitex\_edit/}} directory of the evaluation-code repository and covers glyph-video rendering, inference, the optional Composite post-processing wrapper, and the two-stage training recipe.

\item \textbf{ViTeX-Bench-Leaderboard} (GitHub Pages). \url{https://vitex-bench.github.io/ViTeX-Bench-Leaderboard/}. Submitters attach the \texttt{eval.json} produced by the evaluation code to a submission issue on the leaderboard repository; the maintainers review each submission before adding it to the public leaderboard. Each method is listed with its full 13-metric vector, the primary metrics of \Cref{sec:primary}, and its Pareto-set membership. The leaderboard is pre-populated with all methods reported in \Cref{tab:main}, including the Source video row.
\end{itemize}

\section{Datasheet for ViTeX-Dataset}
\label{app:datasheet}

We follow the Datasheets for Datasets template of Gebru et al.~\cite{datasheets}. The condensed answers below describe \VD; \VB scoring details are provided in \Cref{sec:metrics,sec:baselines}.

\textbf{Motivation.} The dataset was created to support two coupled goals: providing a high-quality paired training set for video scene text editing, and serving as a frozen evaluation benchmark with character-level metrics. It was constructed by the authors and was not sponsored by any commercial entity.

\textbf{Composition.} The dataset contains $387$ source videos. The $230$ training videos are distributed as full $(V, \tilde V, M, s_{\mathrm{src}}, s_{\mathrm{tgt}})$ tuples, where $\tilde V$ is a high-quality paired edited video produced by the construction pipeline; the $157$ evaluation videos are distributed as $(V, M, s_{\mathrm{src}}, s_{\mathrm{tgt}})$ tuples without edited videos. We do not call $\tilde V$ a ground-truth edit because there is no single canonical rendering of the requested string: font, weight, color, and lighting interaction are not uniquely determined by the instruction. All videos are $1280\!\times\!720$ MP4/H.264 videos with $120$ frames at $24$\,fps. Source and target string lengths are closely matched (mean $8.0\!\pm\!5.7$ vs.\ $8.0\!\pm\!5.4$ characters; ranges $1$--$41$ and $1$--$36$). Source videos originate from Panda-70M~\cite{panda70m} and InternVid~\cite{internvid}, both of which are derived from publicly available web videos.

\textbf{Collection.} Source videos were retrieved by an automatic candidate pipeline applied to Panda-70M and InternVid (see \Cref{app:pipeline} for details) and then manually curated. The screening pass inspected $4{,}322$ Panda-70M candidates and retained $628$, and inspected $3{,}768$ InternVid candidates and retained $200$. The released $387$ videos were selected from this accepted pool. Edits were generated through the foundation-model pipeline described in \Cref{sec:pipeline}. A single author-annotator performed the original candidate screening, SAM 3 keypoint prompting, target-string audit, motion classification, and final paired-video selection. An independent annotator subsequently repeated mask annotation for 12 difficulty-stratified clips (\Cref{app:coverage-validation}).

\textbf{Preprocessing.} Source videos are re-encoded to a fixed resolution and frame rate. SAM 3 keypoint masks undergo the fixed morphological dilation described in \Cref{app:pipeline}. Qwen3-VL replacement strings are audited and re-sampled when necessary. The PISCO inserter is fine-tuned under the first-frame-reference protocol with amodal-completion supervision in the text region before being used in the pipeline.

\textbf{Uses.} The dataset is intended for evaluating and training video scene-text editing models. It is not intended for forensic or evidentiary use, personal identification, medical-label modification, real-world license-plate manipulation, or generation of misleading news content.

\textbf{Distribution.} The dataset is released under CC-BY-NC 4.0 for non-commercial research only. The upstream source datasets are non-commercial as well---Panda-70M is distributed under the Snap Inc.\ Non-Commercial Research License, and InternVid is distributed under CC-BY-NC-SA 4.0---and the CC-BY-NC 4.0 release of \VD preserves their non-commercial restriction; users who redistribute InternVid-derived portions remain bound by its additional ShareAlike (CC-BY-NC-SA 4.0) terms. The dataset is hosted on Hugging Face at \url{https://huggingface.co/datasets/ViTeX-Bench/ViTeX-Dataset}.

\textbf{Maintenance.} The maintainers commit to issue tracking and version updates for the first three years; a community steering committee will assume maintenance afterward. Croissant~\cite{croissant} metadata with Responsible AI fields is distributed alongside the dataset.

\section{Pipeline Details}
\label{app:pipeline}

\textbf{Source retrieval.} Panda-70M and InternVid candidates are scored by a hit-tag pipeline that searches each video's caption for nine strong-positive text-rich categories---blackboard/whiteboard, signage, poster/notice, banner/slogan, billboard, license-plate, scoreboard, menu/document, and label/sticker---plus a weak tenth group (\texttt{text|words|letters|writing}) that only counts when paired with a real-world-context regex over people, settings, and surfaces (e.g., person, classroom, store, stadium, jersey, window, paper). A negative pattern set demotes seven failure modes: animation/cartoon, CGI/render, gameplay, synthetic/AI-generated, screen recording, logo/intro, and black-background title cards. Panda-70M candidates additionally clear a video-level matching-score threshold; InternVid candidates pass Aesthetic and UMT score cutoffs. Surviving candidates are downloaded with \texttt{yt-dlp}, deduplicated to one segment per source video, and standardized via \texttt{ffmpeg} (libx264, CRF 18) to $1280\!\times\!720$, $24$\,fps, $120$ frames, with longer InternVid videos center-cropped to a $5$-second window and any video carrying an internal scene cut (FFmpeg \texttt{scdet}, threshold $0.3$) rejected. The author-annotator retained only videos with a clearly readable, edit-suitable text region and no obvious sensitive content.

\textbf{SAM~3 segmentation.} A custom interactive web GUI (FastAPI + browser) wraps the SAM~3 video predictor in float16 AMP and supports per-object multi-target annotation with no cap on prompt-point count. The annotator opens a video and marks the editable glyph region on the first frame with positive keypoints (placed inside the glyph) and negative keypoints (placed on neighboring non-glyph pixels such as background, adjacent objects, or cast shadow that would otherwise leak into the mask); SAM~3 then propagates the resulting binary mask forward to the remaining $119$ frames. The annotator scrubs the propagated mask frame by frame, adds corrective keypoints on the worst-drifting frame, and re-runs propagation until the mask tracks the glyph cleanly across all $120$ frames. The cleaned per-video mask is exported as a binary mask video. Morphological dilation uses a $25\!\times\!25$\,px elliptical structuring element applied for $3$ iterations to give downstream editors a margin beyond the glyph boundary.

\textbf{Qwen3-VL target-text generation.} The vision--language model Qwen3-VL-32B-Instruct is run locally via Ollama (\texttt{qwen3-vl:32b-instruct}, Q4\_K\_M quantization, 32K context) at temperature 0.1. The first frame is cropped to the dilated mask bounding box with a 28\% relative margin and resized to a 1280-px long side. The model is asked to (i) read the source string $s_{\mathrm{src}}$ from the crop and (ii) propose one $s_{\mathrm{tgt}}$ that differs from $s_{\mathrm{src}}$, approximately matches its length, fits the scene semantics, and avoids offensive, political, or trademark content. A repair pass at temperature 0.25 is run when the JSON response is malformed. The author-annotator audits all sampled candidates, and rejected candidates are re-sampled before the video is accepted. Per-record rejection and resampling counts were not logged in the initial release.

\textbf{Removal.} We use removal-1.3B, released with PISCO~\cite{pisco} as a Wan2.1-VACE-1.3B fine-tune with ROSE-style side-effect-aware training; inference uses 50 steps, the released classifier-free-guidance configuration, and the dilated mask $M$.

\textbf{Google Gemini 3 Pro Image first-frame edit.} We use the Google Gemini 3 Pro Image API (also known as Nano Banana Pro)~\cite{nanobananapro} for the first-frame rewrite described in \Cref{sec:pipeline}. The prompt extends the per-video instruction stored alongside each task tuple (e.g., \emph{``Change GAUGES to SCALES; preserve everything else.''}) with a coarse spatial qualifier inferred from the first-frame mask centroid: \texttt{top-left}, \texttt{top-right}, \texttt{center}, \texttt{bottom-left}, \texttt{bottom-right}, \texttt{left side}, or \texttt{right side}. The full prompt thus reads \emph{``Change \texttt{<source>} on the \texttt{<region>} of the picture to \texttt{<target>}; preserve everything else.''} Failed first-frame edits are re-sampled and manually rechecked; videos for which no successful first-frame edit is obtained are discarded. Per-record retry counts were not logged in the initial release.

\textbf{Strategy~A (alpha composition).} Static videos are identified by visual inspection of the source video; the check asks whether the text-region position and shape remain anchored across all $120$ frames. Each static video is composed by both Strategy~A and Strategy~B, and the higher-quality output is retained after inspection. Dynamic videos bypass Strategy~A and use Strategy~B exclusively. In the final paired training split, $56$ videos use Strategy~A and $174$ use Strategy~B.

\textbf{Strategy~B (PISCO inserter fine-tune).} The inserter is the dual-branch PISCO-14B (high-noise + low-noise) further fine-tuned for our setting at 720p$\times$121 frames. Each branch starts from the released PISCO-14B-720p121 base model and is fine-tuned with AdamW, learning rate $5\!\times\!10^{-5}$, 20 warmup steps, and gradient accumulation 8 on $8\!\times\!\text{H100~80GB}$ with DeepSpeed ZeRO-3 and CPU offload (per-GPU peak $\sim$17\,GiB). At inference time, the first-frame patch $p_1^{\text{new}}$ is composited onto frame 1 of $V_{\text{clean}}$ to form a single-keyframe reference video; the same $p_1^{\text{new}}$ alpha is propagated along the original mask trajectory to form the per-frame reference mask. Depth from Depth Anything~3~\cite{depthanything3} provides geometric priors. The fine-tune training set is constructed automatically from a separate corpus of scene-text videos disjoint from the $387$-video release. Applying removal-1.3B to each video yields a (text-removed, original) video pair, which we treat as a (clean-background input, text-inserted target) supervision sample with the original first frame serving as the keyframe reference. No additional human labeling is required for this auxiliary set, and the loss is amodal completion in the text region.

\section{Training and Evaluation Splits}
\label{app:split}

The $157$-video evaluation split is permanently frozen. Its source-text Unicode blocks fall into four scripts: Latin ($150$ videos), Chinese ($4$), Japanese ($1$), and Cyrillic ($2$); the OCR backend selects the recognizer per video from this routing.

\section{Evaluation Metric Implementation}
\label{app:metrics-impl}

\textbf{OCR backend.} PaddleOCR's PP-OCRv5~\cite{ppocrv5} ships separate pretrained detection and recognition checkpoints per language family. We route each video to one recognizer from \{Latin (\texttt{en}), Chinese (\texttt{ch}), Japanese (\texttt{japan}), Cyrillic (\texttt{ru})\} according to the source-text Unicode block; per-script counts are listed in \Cref{app:split}. Recognized boxes below confidence $0.30$ are dropped. Surviving strings are normalized in three steps---NFKC folding, case folding, and whitespace/punctuation removal---so that, for example, \texttt{35,000} and \texttt{35 000} compare as equal characters.

\textbf{Substring edit distance.} $d_{\mathrm{sub}}(r,c)$, also known in sequence alignment as fitting or semi-global edit distance, is the minimum edit distance between reference $r$ and any contiguous substring of candidate $c$. As a worked example, for target \texttt{BIG} inside the longer OCR string \texttt{ABIGA}, standard Levenshtein distance is $2$ because it charges the two surrounding \texttt{A} characters; substring distance is $0$ because \texttt{BIG} appears exactly. We compute $d_{\mathrm{sub}}$ with a Wagner--Fischer dynamic-programming table whose first row is initialized to zero, making prefixes of candidate $c$ free. The first column $\mathrm{dp}[i,0]=i$ accumulates the cost of matching the first $i$ characters of $r$ to an empty candidate substring, and the returned distance is $\min_j \mathrm{dp}[|r|, j]$, making suffixes of $c$ free.

\textbf{Edge cases.} Five of the 157 evaluation videos have no detectable source frames ($\mathcal{D}=\emptyset$); they are excluded from SeqAcc and CharAcc means, leaving 152 supported videos. They remain in metrics that do not require source detectability. TTS is omitted for videos with no adjacent detectable pair ($\mathcal{P}=\emptyset$).

\textbf{Text-crop bounding box.} Each video's crop scope is fixed across the $120$ frames as the union of $\{m_t\}_{t=1}^{T}$ enlarged by a $16$-pixel margin and clipped to the frame border, rather than as a per-frame bounding box. The temporal metrics $\mathrm{Flicker}_c$ and $\mathrm{Warp}_c$ therefore measure glyph drift inside a stationary window rather than bounding-box jitter. In $\mathrm{Warp}_S$, the backward warp $\mathcal{W}$ is sampled at valid RAFT-flow pixels only.

\textbf{Locality composite and PSNR reporting.} The locality composite $\hat f_t^{\mathrm{loc}}$ uses the same dilated mask $M$ that gates the text-region crop. The Source video row compares the decoded source with itself: $\hat f_t^{\mathrm{loc}}=f_t$ exactly, so its $\mathrm{MSE}=0$ and PSNR is mathematically $\infty$. Composite outputs undergo an additional lossy encode/decode cycle and retain small boundary differences, yielding finite PSNR. All other rows have finite video-level PSNR values in the released evaluator.

\textbf{Statistical reporting.} Each evaluation-split aggregate is the video-level mean over its support set, i.e., the videos for which the metric is defined. We report 95\% confidence intervals via percentile bootstrap with 1000 video-level resamples drawn with replacement; the bootstrap RNG is seeded for reproducibility. Per-video metric values, support sizes, and CIs are written into the released evaluation artifacts. \Cref{tab:text-ci,tab:vq-ci,tab:loc-ci} list the per-method bootstrap CIs for all 13 metrics, grouped by axis.

\begin{table}[h]
\caption{Headline text-correctness means with 95\% bootstrap confidence intervals on the $157$-video evaluation split. Text-correctness metrics are scored on source-detectable frames only (\Cref{sec:metrics}); $5$ videos with no detectable source frame are excluded from the means.}
\label{tab:text-ci}
\centering
\footnotesize
\setlength{\tabcolsep}{3pt}
\begin{tabular}{lccc}
\shline
\textbf{Method} & \textbf{SeqAcc} & \textbf{CharAcc} & \textbf{TTS} \\
\hline
Source video & 0.000 [0.000, 0.000] & 0.317 [0.285, 0.350] & 0.760 [0.721, 0.800] \\
\hline
AnyText2 & 0.280 [0.229, 0.333] & 0.633 [0.591, 0.676] & 0.382 [0.336, 0.427] \\
TextCtrl & 0.475 [0.409, 0.539] & 0.734 [0.684, 0.780] & 0.511 [0.459, 0.559] \\
FLUX-Text & 0.528 [0.483, 0.578] & 0.737 [0.696, 0.778] & 0.326 [0.286, 0.369] \\
RS-STE & 0.354 [0.288, 0.412] & 0.626 [0.574, 0.677] & 0.534 [0.484, 0.588] \\
TextCtrl + AnyV2V & 0.057 [0.031, 0.088] & 0.308 [0.274, 0.346] & 0.257 [0.222, 0.297] \\
Wan2.1-VACE-14B & 0.000 [0.000, 0.000] & 0.298 [0.267, 0.332] & 0.689 [0.647, 0.734] \\
VideoPainter & 0.364 [0.300, 0.434] & 0.619 [0.559, 0.676] & 0.606 [0.554, 0.656] \\
Kling Video 3.0 Omni & 0.000 [0.000, 0.000] & 0.208 [0.178, 0.239] & 0.641 [0.594, 0.690] \\
\textbf{\VX} & 0.363 [0.289, 0.436] & 0.714 [0.670, 0.754] & 0.658 [0.609, 0.703] \\
\textbf{\VX\ (text-only)} & 0.339 [0.265, 0.411] & 0.688 [0.644, 0.731] & 0.653 [0.600, 0.706] \\
\textbf{\VX\ (Composite)} & 0.361 [0.287, 0.435] & 0.718 [0.675, 0.758] & 0.664 [0.614, 0.712] \\
\textbf{\VX\ (text-only, Composite)} & 0.347 [0.273, 0.422] & 0.691 [0.650, 0.733] & 0.654 [0.600, 0.706] \\
\shline
\end{tabular}
\end{table}

\begin{table}[h]
\caption{Visual-quality means with 95\% bootstrap confidence intervals on the $157$-video evaluation split. $f$/$c$ denotes full-frame/text-crop scope. Lower is better for Flicker and Warp; higher is better for MUSIQ. VideoPainter Flicker and Warp values are not directly comparable (\Cref{app:baselines}).}
\label{tab:vq-ci}
\centering
\footnotesize
\setlength{\tabcolsep}{2.5pt}
\resizebox{\linewidth}{!}{
\begin{tabular}{lcccccc}
\shline
\textbf{Method} & \textbf{Flicker$_f\downarrow$} & \textbf{Flicker$_c\downarrow$} & \textbf{Warp$_f\downarrow$} & \textbf{Warp$_c\downarrow$} & \textbf{MUSIQ$_f\uparrow$} & \textbf{MUSIQ$_c\uparrow$} \\
\hline
Source video         & 3.72 [3.14, 4.41] & 3.68 [2.91, 4.63] & 1.46 [1.33, 1.62] & 1.27 [1.09, 1.46] & 70.33 [69.55, 71.14] & 45.12 [43.23, 47.09] \\
\hline
AnyText2             & 3.34 [2.89, 3.88] & 4.95 [4.36, 5.64] & 2.04 [1.85, 2.28] & 3.95 [3.56, 4.41] & 66.68 [65.82, 67.57] & 41.65 [39.96, 43.45] \\
TextCtrl             & 3.80 [3.22, 4.49] & 4.29 [3.53, 5.21] & 1.59 [1.44, 1.76] & 2.09 [1.86, 2.35] & 70.32 [69.52, 71.14] & 42.77 [40.96, 44.74] \\
FLUX-Text            & 5.11 [4.50, 5.80] & 14.81 [13.56, 16.01] & 3.03 [2.80, 3.29] & 13.01 [11.71, 14.21] & 70.26 [69.45, 71.07] & 43.85 [42.09, 45.74] \\
RS-STE               & 3.73 [3.15, 4.40] & 3.66 [2.97, 4.53] & 1.61 [1.46, 1.77] & 1.81 [1.60, 2.06] & 69.57 [68.74, 70.43] & 34.26 [32.67, 36.01] \\
TextCtrl + AnyV2V    & 4.98 [4.42, 5.67] & 4.98 [4.17, 5.89] & 4.11 [3.68, 4.59] & 3.97 [3.46, 4.56] & 69.41 [68.41, 70.46] & 33.85 [32.37, 35.35] \\
Wan2.1-VACE-14B      & 3.78 [3.21, 4.44] & 3.84 [3.09, 4.77] & 1.69 [1.53, 1.86] & 1.56 [1.36, 1.79] & 70.54 [69.75, 71.34] & 45.26 [43.38, 47.30] \\
VideoPainter$^{\dagger}$ & 2.38 [2.07, 2.73] & 2.62 [2.19, 3.12] & 2.93 [2.46, 3.47] & 3.35 [2.72, 4.10] & 67.16 [66.22, 68.13] & 40.59 [39.19, 42.09] \\
Kling Video 3.0 Omni & 4.25 [3.59, 5.03] & 4.08 [3.27, 5.05] & 3.12 [2.59, 3.78] & 2.90 [2.30, 3.65] & 72.23 [71.62, 72.85] & 47.75 [45.68, 49.81] \\
\textbf{\VX}                  & 3.35 [2.85, 3.92] & 3.58 [2.88, 4.40] & 1.58 [1.44, 1.74] & 1.62 [1.40, 1.85] & 70.29 [69.48, 71.14] & 46.22 [44.51, 47.95] \\
\textbf{\VX\ (text-only)}      & 3.29 [2.81, 3.84] & 3.45 [2.78, 4.23] & 1.56 [1.42, 1.71] & 1.55 [1.34, 1.77] & 69.81 [68.97, 70.68] & 43.84 [42.11, 45.63] \\
\textbf{\VX\ (Composite)}     & 3.75 [3.16, 4.44] & 3.92 [3.13, 4.87] & 1.52 [1.38, 1.67] & 1.61 [1.41, 1.84] & 70.39 [69.57, 71.20] & 46.84 [45.05, 48.54] \\
\textbf{\VX\ (text-only, Composite)} & 3.74 [3.15, 4.43] & 3.83 [3.07, 4.74] & 1.51 [1.37, 1.67] & 1.56 [1.36, 1.78] & 70.37 [69.56, 71.20] & 45.09 [43.30, 46.93] \\
\shline
\end{tabular}}
\end{table}

\begin{table}[h]
\caption{Edit-locality means with 95\% bootstrap confidence intervals on the $157$-video evaluation split. PSNR/SSIM are higher-is-better; LPIPS/DreamSim are lower-is-better. The Source video row satisfies $\hat f_t = f_t$ exactly, so its PSNR is mathematically $\infty$; LPIPS and DreamSim are exactly $0$. Bounding-box-local Family-A editors (TextCtrl, RS-STE) copy exterior pixels from the source before encoding (\Cref{app:baselines}).}
\label{tab:loc-ci}
\centering
\footnotesize
\setlength{\tabcolsep}{3pt}
\begin{tabular}{lcccc}
\shline
\textbf{Method} & \textbf{PSNR$\uparrow$} & \textbf{SSIM$\uparrow$} & \textbf{LPIPS$\downarrow$} & \textbf{DreamSim$\downarrow$} \\
\hline
Source video         & $\infty$              & 1.000 [1.000, 1.000] & 0.000 [0.000, 0.000] & 0.000 [0.000, 0.000] \\
\hline
AnyText2             & 25.56 [25.11, 26.06] & 0.905 [0.894, 0.915] & 0.091 [0.087, 0.096] & 0.043 [0.040, 0.046] \\
TextCtrl             & 41.14 [40.68, 41.61] & 0.994 [0.994, 0.995] & 0.008 [0.007, 0.009] & 0.004 [0.004, 0.005] \\
FLUX-Text            & 31.49 [31.13, 31.85] & 0.975 [0.973, 0.976] & 0.029 [0.027, 0.030] & 0.012 [0.011, 0.013] \\
RS-STE               & 37.00 [36.71, 37.30] & 0.983 [0.982, 0.984] & 0.024 [0.022, 0.025] & 0.007 [0.007, 0.008] \\
TextCtrl + AnyV2V    & 21.08 [20.67, 21.51] & 0.785 [0.769, 0.801] & 0.225 [0.212, 0.239] & 0.073 [0.066, 0.083] \\
Wan2.1-VACE-14B      & 35.21 [34.75, 35.64] & 0.976 [0.974, 0.978] & 0.022 [0.021, 0.023] & 0.007 [0.006, 0.008] \\
VideoPainter         & 28.56 [28.06, 29.00] & 0.915 [0.905, 0.924] & 0.104 [0.097, 0.112] & 0.024 [0.022, 0.026] \\
Kling Video 3.0 Omni & 21.18 [20.46, 21.97] & 0.843 [0.824, 0.861] & 0.176 [0.156, 0.196] & 0.061 [0.053, 0.069] \\
\textbf{\VX}                  & 29.70 [29.31, 30.07] & 0.953 [0.948, 0.957] & 0.060 [0.056, 0.064] & 0.020 [0.018, 0.021] \\
\textbf{\VX\ (text-only)}      & 29.25 [28.80, 29.66] & 0.953 [0.949, 0.958] & 0.058 [0.055, 0.062] & 0.023 [0.021, 0.025] \\
\textbf{\VX\ (Composite)}     & 42.67 [42.46, 42.88] & 0.992 [0.992, 0.993] & 0.007 [0.007, 0.008] & 0.003 [0.002, 0.003] \\
\textbf{\VX\ (text-only, Composite)} & 42.67 [42.46, 42.87] & 0.992 [0.992, 0.993] & 0.007 [0.007, 0.008] & 0.003 [0.002, 0.003] \\
\shline
\end{tabular}
\end{table}

\section{Baseline Implementation}
\label{app:baselines}

\textbf{Family A --- per-frame image scene-text editing.}
AnyText2~\cite{anytext2}, TextCtrl~\cite{textctrl}, FLUX-Text~\cite{fluxtext}, and RS-STE~\cite{rsste} are each applied zero-shot to every frame using the official pretrained weights released by the authors. Per-frame inference produces 120 independently edited frames that are concatenated into the output video with no temporal coupling. The four editors operate at different native resolutions and resampling levels, summarized in \Cref{tab:famA-config}. AnyText2 (SD-1.5 backbone) ingests the full $1280\!\times\!720$ frame downsampled to its native $1024\!\times\!576$ working resolution and Lanczos-upsamples the output back to $1280\!\times\!720$; FLUX-Text inherits the FLUX backbone's native resolution and similarly processes the full frame. TextCtrl and RS-STE, in contrast, are bounding-box-local: they crop the dilated-mask bounding box from the source frame, run the editor at the model's fixed working resolution ($256\!\times\!256$ for TextCtrl, $32\!\times\!128$ for RS-STE), Lanczos-upsample the output back to the bounding-box size, and alpha-composite it into the original frame using the dilated mask. Pixels outside the bounding box are copied from the source before encoding. This bounding-box-local design has a structural consequence for edit locality: exterior differences are constrained by the paste boundary, resampling, and final video encoding rather than full-frame synthesis. In the decoded evaluation video, lossy compression can also introduce small differences beyond the paste boundary. TextCtrl and RS-STE therefore obtain high PSNR/SSIM and low LPIPS, reflecting direct pixel preservation rather than learned background reconstruction. We retain these raw locality scores because preservation is part of the task, and apply a common Composite wrapper to all editors as a separate control (\Cref{app:composite}). The four representatives span multilingual diffusion, structure/style disentanglement, FLUX-based regional attention, and recognition-supervised editing, providing a varied comparison of per-frame approaches.

\begin{table}[h]
\caption{Family-A per-frame editor working resolutions and the resampling pipeline applied to each $1280\!\times\!720$ source frame. ``Full'' = the editor processes the entire frame; ``bounding box'' = it only modifies pixels inside the dilated-mask bounding box.}
\label{tab:famA-config}
\centering
\small
\setlength{\tabcolsep}{6pt}
\begin{tabularx}{\linewidth}{@{}llcX@{}}
\shline
\textbf{Editor} & \textbf{Scope} & \textbf{Working resolution} & \textbf{Per-frame pipeline} \\
\hline
AnyText2  & full & $1024\!\times\!576$ & $1280\!\times\!720 \to 1024\!\times\!576 \to$ edit $\to$ Lanczos $\to 1280\!\times\!720$ \\
FLUX-Text & full & FLUX native         & $1280\!\times\!720 \to$ FLUX native $\to$ edit $\to$ resample $\to 1280\!\times\!720$ \\
TextCtrl  & bounding box & $256\!\times\!256$  & bounding-box crop $\to 256\!\times\!256 \to$ edit $\to$ Lanczos $\to$ alpha-composite back \\
RS-STE    & bounding box & $32\!\times\!128$   & bounding-box crop $\to 32\!\times\!128 \to$ edit $\to$ Lanczos $\to$ alpha-composite back \\
\shline
\end{tabularx}
\end{table}

\textbf{Family B --- first-frame edit and image-to-video propagation.}
TextCtrl edits the first frame with its official pretrained weights, identical to its Family-A configuration. AnyV2V~\cite{anyv2v} then propagates the edit by injecting temporal features from the edited first frame into a frozen image-to-video backbone (I2VGen-XL~\cite{i2vgenxl}, the default in the official AnyV2V release). The backbone produces output at its native $512\!\times\!512$ spatial resolution; we Lanczos-upsample each output frame from $512\!\times\!512$ to $1280\!\times\!720$ before evaluation. Default tuning-free hyperparameters from the AnyV2V repository are used.

\textbf{Family C --- mask-conditioned video inpainting.}
Wan2.1-VACE-14B~\cite{vace} runs zero-shot at $1280\!\times\!720$ / $24$\,fps with a $121$-frame native output (Wan's causal latent grid produces $4n\!+\!1$ frames at $n=30$); the trailing frame is dropped to align with the $120$-frame evaluation grid. The model receives the dilated text-region mask $M$ and the same fixed prompt template as Family~D. VideoPainter~\cite{videopainter} is built on the CogVideoX~1.0 5B image-to-video backbone, whose latent grid fixes the output at $720\!\times\!480$ spatial / $49$ frames / $8$\,fps. Running it under our protocol therefore requires both temporal and spatial adaptation. The $120$-frame source video at $1280\!\times\!720$ is temporally downsampled to $40$ frames at $8$\,fps by retaining every third frame, then padded with $9$ repetitions of the last frame to reach the required $49$ input frames; the input is also spatially downsampled to $720\!\times\!480$. VideoPainter outputs $49$ frames at $720\!\times\!480$; we drop the $9$ padding frames at the end, linearly blend-interpolate the remaining $40$ frames at $8$\,fps to $120$ frames at $24$\,fps (with the last frame padded if interpolation falls one short), and Lanczos-upsample the spatial dimension to $1280\!\times\!720$. The same fixed prompt template as Family~D is used; no per-video prompt tuning is applied. Linear blend interpolation mechanically reduces adjacent-frame differences and changes motion-compensated residuals, so VideoPainter's $\mathrm{Flicker}_{f/c}$ and $\mathrm{Warp}_{f/c}$ readings are partly artifacts of the adaptation pipeline rather than direct measurements of the underlying inpainter. We report them as-is and mark them with $\dagger$ in \Cref{tab:main}, excluding them from temporal-metric ranking.

\textbf{Family D --- instruction-guided video-to-video editing.}
Kling Video 3.0 Omni~\cite{kling} is queried through its public web interface with a fixed instruction template. Each of the $157$ evaluation videos is uploaded manually, and the returned video (a $1280\!\times\!720$ / $24$\,fps / $121$-frame video) has its trailing frame dropped to align with the $120$-frame evaluation grid; otherwise the web-interface output is used as-is. Product-version and query-date metadata are recorded with the released evaluation artifacts.

\section{ViTeX-Edit-14B Implementation Details}
\label{app:vitex14b-impl}

\textbf{Backbone configuration.} For reproducibility, the Wan2.1-VACE-14B checkpoint uses a DiT hidden dimension of 5120, 40 attention heads, an FFN dimension of 13,824, and a 40-block main DiT trunk. Its VACE branch attaches eight VACE blocks at trunk layers 0, 5, 10, 15, 20, 25, 30, and 35. The VCU input has 96 channels: a 16-channel inactive latent $\mathrm{VAE}(V\odot(1-M))$, a 16-channel reactive latent $\mathrm{VAE}(V\odot M)$, and a 64-channel patch-unfolded binary mask latent. The noised denoising latent $x_t$ remains on the main DiT trunk; in the first VACE block, the embedded VCU tokens are projected and added to the trunk hidden state. Trainable parameters include the VACE blocks, glyph encoder, and per-block condition cross-attention ($\approx$ 4020M, 30\% of the frozen-DiT backbone; the glyph encoder accounts for $\approx$132M, the eight VACE blocks for $\approx$3.89B, and the VACE patch embedding for $\approx$2M).

\textbf{Glyph video construction.} Qwen3-VL~\cite{qwen3vl} inspects the source-text region in the first frame and selects the closest matching typeface from a curated font library for Latin scripts. Non-Latin scripts (CJK, Cyrillic, dingbats and other Unicode symbols) bypass this selection step and use a script-specific default font. The selected typeface is used to render the target string $s_{\mathrm{tgt}}$ as a white-on-black glyph image, super-sampled at $2\times$ the detected text quadrilateral to improve robustness under projective warping. EasyOCR~\cite{easyocr} detects the source text inside the mask in one of the first ten frames; a detected line is trimmed to the characters matching $s_{\mathrm{src}}$, and only detections centered in and mostly inside the mask are kept. CoTracker3~\cite{cotracker3} then tracks the quadrilateral's corners across the clip. Each tracked quadrilateral is checked against that frame's mask (convex, centered, at least $70\%$ inside, and of consistent size relative to the mask). A frame that fails the check, e.g., under tracker drift, zoom, or occlusion, places the quadrilateral from the mask instead, keeping the nearest valid frame's position relative to the mask's rotated rectangle. If fewer than half of the masked frames pass, every frame is placed this way from the detection; if no source text is detected, the glyph fills the mask's rotated rectangle. On the evaluation split, $149$ clips are tracked ($14$ with some mask-placed frames) and $8$ are placed entirely from the mask. Training glyph videos were rendered with first-frame detection and tracking only; the mask checks were added before the corrected evaluation run (\Cref{app:correction}). The framewise quadrilateral drives a projective warp of the rendered glyph image to produce the glyph video $G_{\text{vid}}$, which follows the source text's motion, scale, and perspective while keeping the target string in the selected typeface.

\textbf{Glyph encoder and condition cross-attention.} The pooled glyph token bundle $E_G$ in \Cref{eq:glyph-pool} and the per-block condition cross-attention in \Cref{eq:cond-xattn} are defined in \Cref{sec:method}. The two trainable modules together hold $\approx\!132$M (glyph encoder) and a small per-block residual layer; the patch embedding uses stride $(1,2,2)$ on the Wan-VAE latent $z_G$, the pooling queries are $64$ learnable tokens, and the LayerNorm in both equations is a pre-norm applied to the keys and values (pooling) or to the queries (condition cross-attention).

\textbf{Training schedule.} Stage~1 runs for 5 epochs at 720p$\times$49 frames with AdamW (weight decay 0.01), constant learning rate $5\!\times\!10^{-5}$, effective batch size 64 (8 GPUs $\times$ micro-batch 1 $\times$ gradient accumulation 8), and dataset repeat 10$\times$, requiring $\sim$22\,h on $8\!\times\!\text{H100~80GB}$. Stage~2 performs long-horizon annealing for 2 epochs at 720p$\times$121 frames, learning rate $1\!\times\!10^{-5}$, effective batch size 64, and $\sim$50\,h wall-clock time, initialized from the Stage-1 checkpoint. The loss is Flow-Matching SFT in bf16. System modifications for $8\!\times\!\text{H100}$ feasibility include lazy hint aggregation, CPU offload at block boundaries, gradient checkpointing, and excluding the Wan VAE from ZeRO-3 sharding.

\section{Correction of \VX{} Results and Glyph-Stream Ablation}
\label{app:correction}

\textbf{Correction.} The \VX{} results in the submitted version of this paper were produced with the glyph stream inactive. When loading the fine-tuned checkpoint, the inference code re-initialized the glyph encoder and reset the condition cross-attention layers to their zero-initialized output projections, so the glyph video had no effect on the output; the trained weights were unaffected. We fixed the loader and re-ran inference on all $157$ evaluation clips with the original settings ($1280\!\times\!720$, $121$ frames, $50$ steps, classifier-free guidance scale $5$, seed $42$, bf16) and the glyph renderer described in \Cref{app:vitex14b-impl}. The corrected outputs and their Composite versions are scored with the released evaluation code (PaddleOCR 3.4.1, PaddleX 3.4.3, PaddlePaddle 3.2.2). The glyph-inactive outputs are reported as \VX{} (text-only), the same weights with the glyph stream disabled so that the target string reaches the model only through the text prompt; they are re-scored in the same environment, and their Composite version uses the same wrapper. Elsewhere, \VX{} refers to the corrected outputs, except in the human studies (\Cref{app:human-validation}) and the Composite pipeline check (\Cref{app:composite}), which use the earlier outputs; the supplementary analyses over raw editors in \Cref{app:coverage-validation,app:background-validation} include \VX{} in its glyph configuration only.

\textbf{Glyph-stream ablation.} \Cref{tab:glyph-ablation} compares the corrected run with the glyph-inactive configuration, scored in the same way on the same clips. With the glyph stream, SeqAcc and CharAcc rise by $0.024$ and $0.026$ (paired 95\% CIs $[-0.025, 0.079]$ and $[-0.003, 0.056]$), text-crop MUSIQ by $2.39$, and DreamSim-loc falls by $0.003$, while Flicker$_c$ and Warp$_c$ increase by $0.14$ and $0.07$.

\begin{table}[t]
\centering
\caption{Glyph-stream ablation on the $157$-video evaluation split. \emph{Text-only} runs the same weights with the glyph stream disabled at inference, the configuration behind the submitted results; both rows are scored with the same evaluation code and environment. Means with 95\% bootstrap CIs; $\Delta$ is the paired per-video difference (\VX{} minus text-only) with its paired-bootstrap 95\% CI.}
\label{tab:glyph-ablation}
\small
\setlength{\tabcolsep}{4pt}
\resizebox{\linewidth}{!}{
\begin{tabular}{lccc}
\shline
\textbf{Metric} & \textbf{\VX} & \textbf{Text-only} & \textbf{$\Delta$} \\
\hline
SeqAcc$\uparrow$ & 0.363 [0.289, 0.436] & 0.339 [0.265, 0.411] & $+$0.024 [$-$0.025, $+$0.079] \\
CharAcc$\uparrow$ & 0.714 [0.670, 0.754] & 0.688 [0.644, 0.731] & $+$0.026 [$-$0.003, $+$0.056] \\
TTS$\uparrow$ & 0.658 [0.609, 0.703] & 0.653 [0.600, 0.706] & $+$0.005 [$-$0.033, $+$0.047] \\
Flicker$_f\downarrow$ & 3.35 [2.85, 3.92] & 3.29 [2.81, 3.84] & $+$0.06 [$+$0.03, $+$0.09] \\
Flicker$_c\downarrow$ & 3.58 [2.88, 4.40] & 3.45 [2.78, 4.23] & $+$0.14 [$+$0.09, $+$0.19] \\
Warp$_f\downarrow$ & 1.58 [1.44, 1.74] & 1.56 [1.42, 1.71] & $+$0.02 [$+$0.01, $+$0.04] \\
Warp$_c\downarrow$ & 1.62 [1.40, 1.85] & 1.55 [1.34, 1.77] & $+$0.07 [$+$0.05, $+$0.10] \\
MUSIQ$_f\uparrow$ & 70.29 [69.48, 71.14] & 69.81 [68.97, 70.68] & $+$0.48 [$+$0.39, $+$0.58] \\
MUSIQ$_c\uparrow$ & 46.22 [44.51, 47.95] & 43.84 [42.11, 45.63] & $+$2.39 [$+$1.94, $+$2.93] \\
PSNR-loc$\uparrow$ & 29.70 [29.31, 30.07] & 29.25 [28.80, 29.66] & $+$0.44 [$+$0.25, $+$0.63] \\
SSIM-loc$\uparrow$ & 0.953 [0.948, 0.957] & 0.953 [0.949, 0.958] & $-$0.001 [$-$0.002, $+$0.001] \\
LPIPS-loc$\downarrow$ & 0.060 [0.056, 0.064] & 0.058 [0.055, 0.062] & $+$0.002 [$+$0.000, $+$0.003] \\
DreamSim-loc$\downarrow$ & 0.020 [0.018, 0.021] & 0.023 [0.021, 0.025] & $-$0.003 [$-$0.004, $-$0.002] \\
\shline
\end{tabular}}
\end{table}

\section{Shared Composite Post-Processing}
\label{app:composite}

Composite is a deterministic, training-free wrapper that takes any editor's raw prediction $\hat V$, the source video $V$, and the shared mask $M$ to produce $\hat V^{\mathrm{Composite}}$. The same parameters are used for all eight baselines and \VX{}. For each frame $t$, the recipe has three steps.

\textbf{Color matching.} Let $B_t$ be the band of pixels obtained by dilating the per-frame mask $m_t$ with a $41\!\times\!41$ structuring element and subtracting $m_t$ itself; $B_t$ captures the local scene context around the text region. We convert both $\hat f_t$ and $f_t$ to the CIELAB color space and compute per-channel band statistics $(\hat\mu_c, \hat\sigma_c)$ for the prediction and $(\mu_c, \sigma_c)$ for the source, with $c\in\{L,a,b\}$. The corrected prediction is the Reinhard mean--variance transfer
\begin{equation}
\label{eq:crop-color}
\tilde f_t^{(c)} = \big(\hat f_t^{(c)} - \hat\mu_c\big)\cdot \frac{\sigma_c}{\hat\sigma_c} + \mu_c,
\end{equation}
clipped to $[0,255]$ and converted back to RGB\@. The transfer falls back to no correction when $|B_t| < 100$ pixels.

\textbf{Feathered alpha.} Let $d_{\mathrm{in}}(x)$ and $d_{\mathrm{out}}(x)$ be the Euclidean distance transforms inside and outside $m_t$. The signed distance $\phi_t = d_{\mathrm{in}} - d_{\mathrm{out}}$ is positive inside the mask and negative outside. The composition alpha is centered on the mask boundary with feather width $w=4$ pixels:
\begin{equation}
\alpha_t = \mathrm{clip}\!\left(\tfrac{\phi_t + w/2}{w},\ 0,\ 1\right).
\label{eq:crop-alpha}
\end{equation}
Thus $\alpha=1$ at $w/2$ pixels inside the mask and $\alpha=0$ at $w/2$ pixels outside, with a linear transition across the boundary.

\textbf{Composition.} The output frame is
\begin{equation}
\label{eq:crop-composite}
\hat f_t^{\mathrm{Composite}} = (1-\alpha_t)\,f_t + \alpha_t\,\tilde f_t.
\end{equation}
Frames are encoded with libx264, CRF 18, yuv420p, $24$\,fps to match the evaluation grid. The complete pipeline (released as \texttt{vitex\_edit/composite.py}) processes the $157$ evaluation videos in $\sim 5$\,min on $8$ CPU workers and requires no GPU.

\textbf{Interpretation.} Before encoding, Composite reproduces the source outside a $w/2$-pixel halo around the mask. The feathered boundary and subsequent compression produce the remaining exterior differences. For the fully scored \VX{} pair, SeqAcc changes from $0.363$ to $0.361$, CharAcc from $0.714$ to $0.718$, and DreamSim-loc from $0.020$ to $0.003$. This comparison measures the effect of restoring source background pixels while retaining the editor's synthesized text region.

\textbf{Application to every baseline.} \Cref{tab:composite-all} reports raw and Composite full-frame Flicker and PSNR-loc using the same libx264 CRF 18 encoding setup. Applied to the originally released, glyph-inactive \VX{} outputs (\Cref{app:correction}), the reproduced pipeline matches their released per-clip Composite PSNR values to a median absolute difference of $0.00$ dB at the displayed precision. Across baselines, Composite PSNR-loc approaches $43$ dB and full-frame Flicker approaches the source value $3.72$. The finite locality level reflects this encoding and blending configuration.

\begin{table}[t]
\centering
\caption{Shared Composite control. Flicker is full-frame; PSNR-loc is in dB. The final column contains raw SeqAcc. Baseline post-Composite OCR was checked only on a sample ($|\Delta|\leq0.04$); the fully re-scored ViTeX-Edit-14B pair appears in \Cref{tab:main}. $^{\dagger}$VideoPainter temporal values are unranked.}
\label{tab:composite-all}
\small
\setlength{\tabcolsep}{4pt}
\resizebox{\linewidth}{!}{
\begin{tabular}{lccccc}
\shline
 & \multicolumn{2}{c}{Flicker$_f\downarrow$} & \multicolumn{2}{c}{PSNR-loc$\uparrow$} & \\
\textbf{Method} & Raw & +Composite & Raw & +Composite & Raw SeqAcc \\
\hline
AnyText2 & 3.34 & 3.85 & 25.6 & 42.9 & 0.280 \\
TextCtrl & 3.80 & 3.78 & 41.1 & 43.0 & 0.475 \\
FLUX-Text & 5.11 & 4.47 & 31.5 & 42.9 & 0.528 \\
RS-STE & 3.73 & 3.71 & 37.0 & 43.0 & 0.354 \\
TextCtrl + AnyV2V & 4.98 & 3.77 & 21.1 & 43.0 & 0.057 \\
Wan2.1-VACE-14B & 3.78 & 3.75 & 35.2 & 43.0 & 0.000 \\
VideoPainter$^{\dagger}$ & 2.38 & 3.73 & 28.6 & 42.9 & 0.364 \\
Kling Video 3.0 Omni & 4.25 & 3.74 & 21.2 & 42.9 & 0.000 \\
\VX & 3.35 & 3.75 & 29.70 & 42.67 & 0.363 \\
\VX\ (text-only) & 3.29 & 3.74 & 29.25 & 42.67 & 0.339 \\
\shline
\end{tabular}}
\end{table}

Color transfer, feathering, and re-encoding can affect glyph readability. The sampled baseline OCR checks therefore do not establish unchanged correctness over the full split. A separate Composite Pareto comparison would require complete text and text-crop temporal re-scoring; the present analysis reports only the measurements supported by the available evaluation.

\section{Croissant + Responsible Use}
\label{app:croissant}

The Croissant metadata file (\texttt{vitex.croissant.json}) follows the MLCommons Croissant 1.0 schema. Responsible AI extension fields are populated, including \texttt{license}, \texttt{citeAs}, \texttt{recordedBy}, \texttt{intendedUse}, \texttt{prohibitedUse}, \texttt{safetyConsiderations}, and \texttt{humanLabelers}. A concise Responsible Use Agreement covering the \VX{} weights is distributed alongside the dataset. Future metadata releases will add per-record rejection/resampling counts and first-frame retry histories, which were not logged in v1.

\section{Detailed Related Work}
\label{app:related}

This appendix expands the per-method positioning that \Cref{sec:related} condenses. Methods used as baselines in \Cref{sec:baselines} or as foundation-model components of the construction pipeline (\Cref{sec:pipeline}) and \VX (\Cref{sec:method}) are described there in detail and are not repeated here.

\paragraph{Closest video predecessor.}
STRIVE~\cite{strive} applies a still-image text edit and photometrically propagates it through a video. Its evaluation centers on the edited region. \VB{} extends the evaluation scope to character correctness over time, full-frame temporal quality, and preservation outside the edit, alongside paired training data and a frozen evaluation split.

\paragraph{Direct inspiration for \VX{}'s conditioning pathway.}
GlyphMastero~\cite{glyphmastero} introduces an explicit glyph encoder to provide stroke-level guidance for image text editing. \VX{} adapts this idea to video: target glyphs follow the source-text quadrilateral across frames, and a learnable encoder supplies the resulting tokens to every VACE block (\Cref{sec:method}). This couples character structure with source-aligned motion.

\paragraph{Concurrent text-related video legibility work.}
VidTextPres~\cite{vidtextpres} studies character preservation in text-to-video generation, while LegiT~\cite{legit} evaluates text legibility in user-generated media. These tasks address readable text in video and media more broadly. \VB{} evaluates a specified replacement string within an existing scene, where text correctness must be balanced with source-motion and background preservation.

\paragraph{Closest evaluation suites.}
EditBoard~\cite{editboard}, FiVE~\cite{five}, IVEBench~\cite{ivebench}, and VEFX-Bench~\cite{vefx} assess instruction-guided video editing along multiple axes. VE-Bench~\cite{vebench} combines human scores with a learned quality predictor; OpenVE-3M~\cite{openve} provides large-scale editing pairs and human ratings; and TDVE-Assessor~\cite{tdvea} studies multimodal quality assessment. \VB{} specializes this evaluation framework to exact text replacement, adding frame-level OCR and decoded-string stability. VBench~\cite{vbench} and VBench-2.0~\cite{vbench2} provide a complementary precedent for decomposed evaluation of video generation. Physics-IQ~\cite{physiq} evaluates physical consistency in generation, the Physics-Aware Video Instance Removal Benchmark~\cite{li2026physics} examines removal of objects and their physical side effects, including shadows and reflections, and PhyFPS-Bench~\cite{pulseofmotion} measures whether generated motion follows a consistent physical time scale. These benchmarks illustrate how task-specific requirements motivate specialized evaluation.

\paragraph{Tuning-free video editing not used as baselines.}
Tuning-free diffusion-based video editors~\cite{fatezero,tokenflow,pix2video,controlvideo,videop2p,rave,slicedit,flowvid,ccedit,motiondirector} use attention or feature propagation to preserve video structure during editing;~\cite{videoeditsurvey} surveys this literature. We include TextCtrl+AnyV2V as a first-frame propagation baseline. Extending the comparison to other systems requires adapting their conditioning and temporal interfaces to the target-string task and the 120-frame evaluation grid.

\paragraph{Image scene-text editing not used as baselines.}
AnyText2~\cite{anytext2}, TextCtrl~\cite{textctrl}, FLUX-Text~\cite{fluxtext}, and RS-STE~\cite{rsste} represent multilingual attribute conditioning, structure/style control, FLUX-based editing, and recognition-supervised editing. Earlier GAN-based~\cite{srnet,swaptext,mostel} and diffusion-based methods~\cite{glyphdraw,diffste,diffute,udifftext,textdiffuser,textdiffuser2,anytext,textflux,textmastero} establish the broader design lineage. Applying an image editor independently to each frame provides no explicit temporal coupling, which helps explain the instability observed in our experiments. Performance of additional image editors remains an empirical question.

\section{Coverage, Ranking Stability, and Annotation Reliability}
\label{app:coverage-validation}

\textbf{Evaluation-size sensitivity.} Of the 157 evaluation clips, 152 contain source-detectable frames and contribute to SeqAcc and CharAcc. We resample these clips with replacement at $n\in\{40,80,120,152\}$, using $B=1{,}000$ replicates and seed $2064$. For each replicate, we compare the nine raw editors' mean-SeqAcc ranking with the full-split ranking and record whether the leading method is retained. The Source anchor and Composite control are excluded.

\begin{table}[t]
\centering
\caption{SeqAcc ranking stability under video-level resampling of the 152 supported evaluation clips. Kendall $\tau$ compares each replicate with the full-split ranking; the final column gives the fraction retaining the leading method.}
\label{tab:rank-stability}
\small
\begin{tabular}{rcc}
\shline
\textbf{Resample size} & \textbf{Mean Kendall $\tau$} & \textbf{Top method retained} \\
\hline
40  & 0.888 & 0.79 \\
80  & 0.910 & 0.89 \\
120 & 0.920 & 0.92 \\
152 & 0.925 & 0.95 \\
\shline
\end{tabular}
\end{table}

Ranking stability increases with sample size. At $n=152$, mean Kendall $\tau$ reaches $0.925$, with the leading method retained in $95\%$ of replicates. Mid-table SeqAcc estimates remain close: RS-STE, \VX{}, and VideoPainter score $0.354$, $0.363$, and $0.364$, respectively, with overlapping intervals (\Cref{tab:text-ci}). The analysis supports the broad ordering within this split while leaving uncertainty about nearby methods. It does not assess coverage beyond the sampled distribution.

\textbf{Difficulty and typography.} The audit of all 387 clips identifies printed ($23\%$), handwritten ($44\%$), and artistic ($33\%$) text. The coverage analysis reports mask-area ratio $0.032\pm0.022$ and approximately $25\%$ static versus $75\%$ dynamic videos. Motion categories are assigned visually; construction-strategy counts differ because static videos can use either strategy. In the four-cell mask-area-by-motion analysis, mean SeqAcc across the nine raw editors ranges from $0.341$ for small-area/static clips ($n=17$) to $0.246$ for large-area/static clips ($n=22$). The contrast between these two static groups indicates variation associated with mask area; it does not isolate a causal effect of area or motion.

\textbf{Non-Latin slice.} Alongside 150 Latin-script clips, the evaluation split contains Chinese ($4$), Japanese ($1$), and Cyrillic ($2$) examples. The OCR and glyph pipelines route these inputs to script-specific recognizers and fonts. On the pooled seven-clip slice, AnyText2 leads with SeqAcc $0.168$ and CharAcc $0.295$; the other eight raw editors have CharAcc below $0.19$. Given the small sample, we report this pooled diagnostic without per-script conclusions or confidence intervals.

\textbf{Independent mask re-annotation.} A second annotator re-labeled 12 difficulty-stratified clips using the same prompting, SAM 3 propagation, correction, and three-iteration $25\!\times\!25$ dilation procedure. On the dilated masks consumed by evaluation, agreement is IoU $0.95$, Dice $0.98$, and crop-box IoU $0.94$. Re-scoring the editor outputs with these masks yields Kendall $\tau=0.89$ for DreamSim-loc and $\tau=1.00$ for text-crop Warp. Absolute changes in method-mean DreamSim-loc have median $0.007$ and maximum $0.051$; the two ordering changes exchange \VX{} with VideoPainter and the two methods with the poorest locality. The rankings are therefore more stable than the absolute scores on this subset.

This audit measures reproducibility under a shared model-assisted pipeline. It covers mask annotation, whereas candidate screening, target selection, motion labels, and final edit acceptance have not received independent agreement studies. Source-string OCR CharAcc $0.966$ is an automatic consistency check, separate from human annotation agreement.

\textbf{Provenance.} Screening retained 628 of 4,322 Panda-70M candidates and 200 of 3,768 InternVid candidates; the final 387 standardized videos were selected from this pool. \Cref{app:pipeline} describes the components and resulting assets. Per-record target-string rejection/resampling counts and first-frame retry histories were not logged in v1, so the aggregate funnel cannot recover those rates. Subsequent metadata releases are planned to record them directly.

\section{OCR Calibration and Human Evaluation}
\label{app:human-validation}

\textbf{Empirical OCR reference levels.} Calibration uses the benchmark's recognizer, normalization, fitting edit distance, and source-detectability gate. We compare source OCR with $s_{\mathrm{src}}$; the Source row in \Cref{tab:main} instead compares it with $s_{\mathrm{tgt}}$. Thus calibration exact match $0.851$ measures recognition of the existing text, whereas Source SeqAcc $0.000$ measures the absence of the requested edit. TTS depends only on adjacent decoded strings and is unchanged by the reference choice.

\begin{table}[t]
\centering
\caption{Empirical OCR reference levels. Source correctness is evaluated against $s_{\mathrm{src}}$ on detectable frames; pipeline-rendered edits are evaluated against $s_{\mathrm{tgt}}$.}
\label{tab:ocr-calibration}
\small
\begin{tabularx}{\linewidth}{@{}lX@{}}
\shline
\textbf{Quantity} & \textbf{Observed value} \\
\hline
Source exact match within $\mathcal D$ & $0.851$ \\
Source CharAcc within $\mathcal D$ & $0.966$ \\
Source decoded-string TTS & $0.760$ \\
Source detectability $|\mathcal D|/T$ & Median $1.00$; 25th percentile $0.97$; mean $0.90$ \\
Clips with $\mathcal D=\emptyset$ & $5$ of $157$ evaluation clips \\
Pipeline-rendered $\tilde V$ & Exact match $0.585$; CharAcc $0.790$; TTS $0.768$ \\
\shline
\end{tabularx}
\end{table}

Source exact match below one and TTS $0.760$ reveal recognition errors and decoded-string instability without editing. Because target strings and renderings can differ from the source, these values provide calibration context rather than universal score bounds. We retain the original metric scale and interpret the calibration on its source-detectable support.

Pipeline-rendered $\tilde V$ yields CharAcc $0.790$, exact match $0.585$, and TTS $0.768$, reflecting both recognition error and possible rendering defects. Human transcription of the rendered-text calibration crops reaches CharAcc $0.917$, with $97\%$ judged readable. The higher human read-back accuracy indicates that OCR can underestimate the readability of generated text, although neither measure establishes error-free rendering.

\textbf{Method-blinded transcription.} One author transcribed 351 method-blinded output crops spanning nine raw editors. The Latin-script calibration excluded nine non-Latin crops and used the same normalization and fitting distance as the automatic evaluation. Embedded real-text catch trials yielded $32/36$ exact transcriptions ($0.889$). Human and OCR method-level CharAcc rankings agree at Spearman $\rho=0.95$, with absolute score differences of $0.00$--$0.13$. This supports similar method ordering on the sampled material despite shifts in absolute accuracy. The study is limited to a single author-annotator.

\textbf{Three-axis rating study.} Three non-author raters each scored the same 70 video outputs, stratified across the nine raw editors, on text correctness, temporal quality, and edit locality. Scores range from 1 to 3, with higher values indicating better quality. We compute ordinal Krippendorff $\alpha$ across raters and Spearman $\rho$ between the mean rating for each output and its automatic score. Both human studies predate the correction in \Cref{app:correction} and use the originally released, glyph-inactive \VX{} outputs.

\begin{table}[t]
\centering
\caption{Human agreement and metric alignment for 70 outputs rated by three non-author raters. Human scores increase with quality; the signs of $\rho$ follow the automatic metrics' directions. All three correlations have $p<0.001$.}
\label{tab:human-alignment}
\small
\begin{tabular}{lclc}
\shline
\textbf{Axis} & \textbf{Ordinal $\alpha$} & \textbf{Automatic metric} & \textbf{Spearman $\rho$} \\
\hline
Text correctness & 0.87 & SeqAcc$\uparrow$ & $+0.71$ \\
Temporal quality & 0.80 & Warp$_c\downarrow$ & $-0.40$ \\
Edit locality & 0.37 & DreamSim-loc$\downarrow$ & $-0.53$ \\
\shline
\end{tabular}
\end{table}

Text-crop Warp correlates more strongly with temporal ratings than full-frame Warp ($\rho=-0.40$ vs.\ $-0.20$), supporting its use as the temporal primary. Text and temporal ratings show substantially higher inter-rater agreement than locality ($\alpha=0.87$, $0.80$, and $0.37$). The locality correlation is therefore interpreted cautiously. These results characterize alignment on the sampled outputs; broader validation would require more raters, scripts, and editing conditions.

\section{Primary-Metric Pareto Comparison}
\label{app:pareto}

\Cref{tab:pareto} applies the dominance rule in \Cref{sec:primary} to raw-output SeqAcc, text-crop Warp, and DreamSim-loc. We exclude the Source anchor, the shared Composite control, and VideoPainter's temporally adapted outputs. The front describes trade-offs among mean scores; confidence intervals in \Cref{tab:text-ci,tab:vq-ci,tab:loc-ci} quantify uncertainty in the underlying estimates.

\begin{table}[t]
\centering
\caption{Raw-output primary metrics and Pareto membership. The comparison uses mean scores; Source, Composite, and temporally adapted VideoPainter outputs are excluded.}
\label{tab:pareto}
\small
\begin{tabular}{lcccc}
\shline
\textbf{Method} & \textbf{On front} & \textbf{SeqAcc$\uparrow$} & \textbf{Warp$_c\downarrow$} & \textbf{DreamSim-loc$\downarrow$} \\
\hline
FLUX-Text & Yes & 0.528 & 13.010 & 0.012 \\
TextCtrl & Yes & 0.475 & 2.088 & 0.004 \\
ViTeX-Edit-14B & Yes & 0.363 & 1.617 & 0.020 \\
RS-STE & Yes & 0.354 & 1.815 & 0.007 \\
ViTeX-Edit-14B (text-only) & Yes & 0.339 & 1.546 & 0.023 \\
Wan2.1-VACE-14B & Yes & 0.000 & 1.561 & 0.007 \\
AnyText2 & No & 0.280 & 3.952 & 0.043 \\
Kling Video 3.0 Omni & No & 0.000 & 2.902 & 0.061 \\
TextCtrl + AnyV2V & No & 0.057 & 3.967 & 0.073 \\
\shline
\end{tabular}
\end{table}

The front exposes distinct operating points. FLUX-Text leads SeqAcc at the cost of large text-crop Warp; TextCtrl combines correctness and locality; the two \VX{} configurations trade SeqAcc against Warp, the text-only one attaining the lowest Warp; and RS-STE pairs lower Warp than TextCtrl with strong locality. Wan2.1-VACE-14B remains non-dominated despite SeqAcc $0$, because stability and locality can be retained without rendering the target. Pareto membership must therefore be interpreted alongside the actual metric values.

\section{Supplementary Background and Identity Diagnostics}
\label{app:background-validation}

We complement the four framewise locality metrics with three probes of background motion, temporal feature stability, and face preservation. These diagnostics are reported separately from the 13 core metrics.

\begin{table}[t]
\centering
\caption{Supplementary preservation diagnostics. BG-Warp and DINOv2-drift use the 157-clip evaluation split; ArcFace-id uses the 131-clip source-face subset. Source is an unranked sanity anchor. $^{\dagger}$VideoPainter's interpolated temporal scores are also unranked.}
\label{tab:background-probes}
\small
\begin{tabular}{lccc}
\shline
\textbf{Method} & \textbf{BG-Warp$\downarrow$} & \textbf{DINOv2-drift$\downarrow$} & \textbf{ArcFace-id$\uparrow$} \\
\hline
Source & 1.67 & 0.0036 & 1.00 \\
TextCtrl & 1.70 & 0.0038 & 0.99 \\
RS-STE & 1.76 & 0.0038 & 0.97 \\
ViTeX-Edit-14B & 1.76 & 0.0036 & 0.87 \\
Wan2.1-VACE-14B & 1.88 & 0.0039 & 0.96 \\
AnyText2 & 2.04 & 0.0046 & 0.82 \\
FLUX-Text & 2.40 & 0.0057 & 0.96 \\
VideoPainter$^{\dagger}$ & 3.05 & 0.0068 & 0.82 \\
Kling Video 3.0 Omni & 3.31 & 0.0044 & 0.69 \\
TextCtrl + AnyV2V & 4.27 & 0.0093 & 0.40 \\
\shline
\end{tabular}
\end{table}

\textbf{Background Warp.} BG-Warp applies the source-flow RAFT error in \Cref{eq:visual} to valid warp pixels outside the mask, $1-m_t$. It measures background residuals after compensation for source motion and inherits the core Warp metric's sensitivity to smoothing and interpolation.

\textbf{Background feature drift.} We extract DINOv2 ViT-L/14 features~\cite{dinov2}, exclude text-mask patches from spatial pooling, and measure one minus the mean cosine similarity between the pooled features of adjacent output frames. If $z_t^{\mathrm{bg}}$ denotes the pooled background feature at frame $t$, the diagnostic is
\begin{equation}
\mathrm{DINOv2\text{-}drift}
=1-\operatorname*{mean}_{t=1}^{T-1}
\cos\!\left(z_t^{\mathrm{bg}},z_{t+1}^{\mathrm{bg}}\right).
\label{eq:bg-drift}
\end{equation}
Low drift indicates stable adjacent-frame features. Natural motion can increase drift, while a consistently altered background can have low drift; source-referenced locality and BG-Warp provide complementary context.

\textbf{Face preservation.} RetinaFace~\cite{retinaface} detects source faces in 131 of the 157 evaluation clips; detection overlays were spot-checked on sampled clips. On this subset, ArcFace-id~\cite{arcface} measures cosine similarity between source and output face embeddings. Source self-comparison gives $1.00$. The probe measures preservation of detectable faces and depends on visibility, detection, and source/output correspondence.

TextCtrl, RS-STE, and \VX{} remain close to the Source anchor on background motion and feature drift. Face similarity separates them: \VX{} scores $0.87$, below TextCtrl ($0.99$), RS-STE ($0.97$), and both FLUX-Text and Wan2.1-VACE-14B ($0.96$). Kling likewise combines relatively low feature drift ($0.0044$) with lower source-face similarity ($0.69$). These contrasts show why temporal stability and source identity require distinct measurements.

Cross-method associations with the core locality metrics are approximately $|\rho|=0.87$ for BG-Warp, $0.65$ for DINOv2 drift, and $0.93$--$0.98$ for ArcFace-id. Background feature drift offers the least redundant signal in this comparison. The correlations indicate shared information on the evaluated methods while leaving room for complementary preservation diagnostics.

%% file: main.bbl
\begin{thebibliography}{10}
\providecommand{\url}[1]{#1}
\csname url@samestyle\endcsname
\providecommand{\newblock}{\relax}
\providecommand{\bibinfo}[2]{#2}
\providecommand{\BIBentrySTDinterwordspacing}{\spaceskip=0pt\relax}
\providecommand{\BIBentryALTinterwordstretchfactor}{4}
\providecommand{\BIBentryALTinterwordspacing}{\spaceskip=\fontdimen2\font plus
\BIBentryALTinterwordstretchfactor\fontdimen3\font minus
  \fontdimen4\font\relax}
\providecommand{\BIBforeignlanguage}[2]{{%
\expandafter\ifx\csname l@#1\endcsname\relax
\typeout{** WARNING: IEEEtran.bst: No hyphenation pattern has been}%
\typeout{** loaded for the language `#1'. Using the pattern for}%
\typeout{** the default language instead.}%
\else
\language=\csname l@#1\endcsname
\fi
#2}}
\providecommand{\BIBdecl}{\relax}
\BIBdecl

\bibitem{hunyuanvideo}
W.~Kong, Q.~Tian, Z.~Zhang \emph{et~al.}, ``{HunyuanVideo}: A systematic
  framework for large video generative models,'' \emph{arXiv preprint
  arXiv:2412.03603}, 2024.

\bibitem{cogvideox}
Z.~Yang, J.~Teng, W.~Zheng, M.~Ding, S.~Huang, J.~Xu, Y.~Yang, W.~Hong,
  X.~Zhang, G.~Feng, D.~Yin, Y.~Zhang, W.~Wang, Y.~Cheng, B.~Xu, X.~Gu,
  Y.~Dong, and J.~Tang, ``{CogVideoX}: Text-to-video diffusion models with an
  expert transformer,'' in \emph{ICLR}, 2025.

\bibitem{ltxvideo}
Y.~HaCohen, N.~Chiprut, B.~Brazowski, D.~Shalem, D.~Moshe, E.~Richardson,
  E.~Levin, G.~Shiran, N.~Zabari, O.~Gordon, P.~Panet, S.~Weissbuch,
  V.~Kulikov, Y.~Bitterman, Z.~Melumian, and O.~Bibi, ``{LTX-Video}: Realtime
  video latent diffusion,'' \emph{arXiv preprint arXiv:2501.00103}, 2025.

\bibitem{wan}
{Team Wan} \emph{et~al.}, ``Wan: Open and advanced large-scale video generative
  models,'' \emph{arXiv preprint arXiv:2503.20314}, 2025.

\bibitem{vace}
Z.~Jiang, Z.~Han, C.~Mao, J.~Zhang, Y.~Pan, and Y.~Liu, ``{VACE}: All-in-one
  video creation and editing,'' in \emph{ICCV}, 2025.

\bibitem{kling}
{Kuaishou Kling Team}, ``{Kling Video 3.0 Omni},''
  \href{https://ir.kuaishou.com/news-releases/news-release-details/kling-ai-launches-30-model-ushering-era-where-everyone-can-be/}{Kuaishou
  press release}, 2026, closed-source commercial reference-based video-to-video
  editor.

\bibitem{moviegen}
A.~Polyak, A.~Zohar, A.~Brown, A.~Tjandra, A.~Sinha, A.~Lee, A.~Vyas, B.~Shi,
  C.-Y. Ma, C.-Y. Chuang \emph{et~al.}, ``Movie gen: A cast of media foundation
  models,'' \emph{arXiv preprint arXiv:2410.13720}, 2024.

\bibitem{sora}
{OpenAI}, ``Sora: Video generation models as world simulators,'' Technical
  report,
  \url{https://openai.com/index/video-generation-models-as-world-simulators/},
  2024.

\bibitem{anytext2}
Y.~Tuo, Y.~Geng, and L.~Bo, ``{AnyText2}: Visual text generation and editing
  with customizable attributes,'' \emph{arXiv preprint arXiv:2411.15245}, 2024.

\bibitem{textctrl}
W.~Zeng, Y.~Shu, Z.~Li, D.~Yang, and Y.~Zhou, ``{TextCtrl}: Diffusion-based
  scene text editing with prior guidance control,'' in \emph{NeurIPS}, 2024.

\bibitem{fluxtext}
R.~Lan, Y.~Bai, X.~Duan, M.~Li, D.~Jin, R.~Xu, D.~Nie, L.~Sun, and X.~Chu,
  ``{FLUX-Text}: A simple and advanced diffusion transformer baseline for scene
  text editing,'' \emph{arXiv preprint arXiv:2505.03329}, 2025.

\bibitem{rsste}
Z.~Fang, P.~Lyu, J.~Wu, C.~Zhang, J.~Yu, G.~Lu, and W.~Pei,
  ``Recognition-synergistic scene text editing,'' in \emph{CVPR}, 2025.

\bibitem{anyv2v}
M.~Ku, C.~Wei, W.~Ren, H.~Yang, and W.~Chen, ``{AnyV2V}: A tuning-free
  framework for any video-to-video editing tasks,'' \emph{Transactions on
  Machine Learning Research (TMLR)}, 2024.

\bibitem{i2vgenxl}
S.~Zhang, J.~Wang, Y.~Zhang, K.~Zhao, H.~Yuan, Z.~Qin, X.~Wang, D.~Zhao, and
  J.~Zhou, ``{I2VGen-XL}: High-quality image-to-video synthesis via cascaded
  diffusion models,'' \emph{arXiv preprint arXiv:2311.04145}, 2023.

\bibitem{videopainter}
Y.~Bian, Z.~Zhang, X.~Ju, M.~Cao, L.~Xie, Y.~Shan, and Q.~Xu, ``{VideoPainter}:
  Any-length video inpainting and editing with plug-and-play context control,''
  in \emph{ACM SIGGRAPH}, 2025.

\bibitem{figedit}
\BIBentryALTinterwordspacing
S.~Li, R.~Rossi, S.~Kim, S.~Choudhary, F.~Dernoncourt, P.~Mathur, Z.~Tu, and
  Y.~Zhao, ``Charts are not images: On the challenges of scientific chart
  editing,'' in \emph{International Conference on Learning Representations},
  2026. [Online]. Available: \url{https://arxiv.org/abs/2512.00752}
\BIBentrySTDinterwordspacing

\bibitem{srnet}
L.~Wu, C.~Zhang, J.~Liu, J.~Han, J.~Liu, E.~Ding, and X.~Bai, ``Editing text in
  the wild,'' in \emph{ACM Multimedia}, 2019.

\bibitem{glyphmastero}
T.~Wang, T.~Liu, X.~Qu, C.~Wu, L.~Liu, and X.~Hu, ``{GlyphMastero}: A glyph
  encoder for high-fidelity scene text editing,'' in \emph{CVPR}, 2025.

\bibitem{textdiffuser}
J.~Chen, Y.~Huang, T.~Lv, L.~Cui, Q.~Chen, and F.~Wei, ``{TextDiffuser}:
  Diffusion models as text painters,'' in \emph{NeurIPS}, 2023.

\bibitem{anytext}
Y.~Tuo, W.~Xiang, J.-Y. He, Y.~Geng, and X.~Xie, ``{AnyText}: Multilingual
  visual text generation and editing,'' in \emph{ICLR}, 2024.

\bibitem{vbench}
Z.~Huang, Y.~He, J.~Yu, F.~Zhang, C.~Si, Y.~Jiang, Y.~Zhang, T.~Wu, Q.~Jin,
  N.~Chanpaisit \emph{et~al.}, ``{VBench}: Comprehensive benchmark suite for
  video generative models,'' in \emph{CVPR}, 2024.

\bibitem{vbench2}
D.~Zheng, Z.~Huang, H.~Liu, K.~Zou, Y.~He, F.~Zhang, L.~Gu, Y.~Zhang, J.~He,
  W.-S. Zheng, Y.~Qiao, and Z.~Liu, ``{VBench-2.0}: Advancing video generation
  benchmark suite for intrinsic faithfulness,'' \emph{arXiv preprint
  arXiv:2503.21755}, 2025.

\bibitem{editboard}
Y.~Chen, P.~Chen, X.~Zhang, Y.~Huang, and Q.~Xie, ``{EditBoard}: Towards a
  comprehensive evaluation benchmark for text-based video editing models,'' in
  \emph{AAAI}, 2025.

\bibitem{five}
M.~Li, C.~Xie, Y.~Wu, L.~Zhang, and M.~Wang, ``{FiVE-Bench}: A fine-grained
  video editing benchmark for evaluating emerging diffusion and rectified flow
  models,'' in \emph{ICCV}, 2025.

\bibitem{ivebench}
Y.~Chen, J.~Zhang, T.~Hu, Y.~Zeng, Z.~Xue, Q.~He, C.~Wang, Y.~Liu, X.~Hu, and
  S.~Yan, ``{IVEBench}: Modern benchmark suite for instruction-guided video
  editing assessment,'' \emph{arXiv preprint arXiv:2510.11647}, 2025.

\bibitem{vebench}
S.~Sun, X.~Liang, S.~Fan, W.~Gao, and W.~Gao, ``{VE-Bench}: Subjective-aligned
  benchmark suite for text-driven video editing quality assessment,''
  \emph{arXiv preprint arXiv:2408.11481}, 2024.

\bibitem{vefx}
X.~Gao, S.~Jiang, B.~Liu, X.~Chen, M.~Yang, S.~Yang, M.~Wu, J.~Yu, Q.~Zheng,
  H.~Wang, J.~Zhang, J.~Yang, Z.~Wang, Q.~Yin, and Z.~Tu, ``{VEFX-Bench}: A
  holistic benchmark for generic video editing and visual effects,''
  \emph{arXiv preprint arXiv:2604.16272}, 2026.

\bibitem{li2026physics}
Z.~Li, X.~Chen, L.~Jiang, D.~Hou, F.~Lin, K.~Yamada, X.~Gao, and Z.~Tu,
  ``Physics-aware video instance removal benchmark,'' \emph{arXiv preprint
  arXiv:2604.05898}, 2026.

\bibitem{panda70m}
T.-S. Chen, A.~Siarohin, W.~Menapace, E.~Deyneka, H.-w. Chao, B.~E. Jeon,
  Y.~Fang, H.-Y. Lee, J.~Ren, M.-H. Yang, and S.~Tulyakov, ``{Panda-70M}:
  Captioning 70m videos with multiple cross-modality teachers,'' in
  \emph{CVPR}, 2024.

\bibitem{internvid}
Y.~Wang, Y.~He, Y.~Li, K.~Li, J.~Yu, X.~Ma, X.~Li, G.~Chen, X.~Chen, Y.~Wang,
  P.~Luo, Z.~Liu, Y.~Wang, L.~Wang, and Y.~Qiao, ``{InternVid}: A large-scale
  video-text dataset for multimodal understanding and generation,'' in
  \emph{ICLR}, 2024.

\bibitem{vdm}
J.~Ho, T.~Salimans, A.~Gritsenko, W.~Chan, M.~Norouzi, and D.~J. Fleet, ``Video
  diffusion models,'' in \emph{Advances in Neural Information Processing
  Systems (NeurIPS)}, 2022.

\bibitem{makeavideo}
U.~Singer, A.~Polyak, T.~Hayes, X.~Yin, J.~An, S.~Zhang, Q.~Hu, H.~Yang,
  O.~Ashual, O.~Gafni \emph{et~al.}, ``Make-a-video: Text-to-video generation
  without text-video data,'' \emph{arXiv preprint arXiv:2209.14792}, 2022.

\bibitem{imagenvideo}
J.~Ho, W.~Chan, C.~Saharia, J.~Whang, R.~Gao, A.~Gritsenko, D.~P. Kingma,
  B.~Poole, M.~Norouzi, D.~J. Fleet, and T.~Salimans, ``Imagen video: High
  definition video generation with diffusion models,'' \emph{arXiv preprint
  arXiv:2210.02303}, 2022.

\bibitem{videoldm}
A.~Blattmann, R.~Rombach, H.~Ling, T.~Dockhorn, S.~W. Kim, S.~Fidler, and
  K.~Kreis, ``Align your latents: High-resolution video synthesis with latent
  diffusion models,'' in \emph{CVPR}, 2023.

\bibitem{animatediff}
Y.~Guo, C.~Yang, A.~Rao, Z.~Liang, Y.~Wang, Y.~Qiao, M.~Agrawala, D.~Lin, and
  B.~Dai, ``{AnimateDiff}: Animate your personalized text-to-image diffusion
  models without specific tuning,'' in \emph{ICLR}, 2024.

\bibitem{opensora2}
Z.~Zheng, X.~Peng, Y.~Lou, C.~Shen, T.~Young, X.~Guo, B.~Wang, H.~Xu, H.~Liu,
  M.~Jiang, W.~Li \emph{et~al.}, ``Open-sora 2.0: Training a commercial-level
  video generation model in {\$200k},'' \emph{arXiv preprint arXiv:2503.09642},
  2025.

\bibitem{tunevideo}
J.~Z. Wu, Y.~Ge, X.~Wang, W.~Lei, Y.~Gu, Y.~Shi, W.~Hsu, Y.~Shan, X.~Qie, and
  M.~Z. Shou, ``Tune-a-video: One-shot tuning of image diffusion models for
  text-to-video generation,'' in \emph{ICCV}, 2023.

\bibitem{fatezero}
C.~Qi, X.~Cun, Y.~Zhang, C.~Lei, X.~Wang, Y.~Shan, and Q.~Chen, ``{FateZero}:
  Fusing attentions for zero-shot text-based video editing,'' in \emph{ICCV},
  2023.

\bibitem{tokenflow}
M.~Geyer, O.~Bar-Tal, S.~Bagon, and T.~Dekel, ``{TokenFlow}: Consistent
  diffusion features for consistent video editing,'' in \emph{ICLR}, 2024.

\bibitem{pix2video}
D.~Ceylan, C.-H.~P. Huang, and N.~J. Mitra, ``{Pix2Video}: Video editing using
  image diffusion,'' in \emph{ICCV}, 2023.

\bibitem{controlvideo}
Y.~Zhang, Y.~Wei, D.~Jiang, X.~Zhang, W.~Zuo, and Q.~Tian, ``{ControlVideo}:
  Training-free controllable text-to-video generation,'' \emph{arXiv preprint
  arXiv:2305.13077}, 2023.

\bibitem{videop2p}
S.~Liu, Y.~Zhang, W.~Li, Z.~Lin, and J.~Jia, ``{Video-P2P}: Video editing with
  cross-attention control,'' in \emph{CVPR}, 2024.

\bibitem{rave}
O.~Kara, B.~Kurtkaya, H.~Yesiltepe, J.~M. Rehg, and P.~Yanardag, ``{RAVE}:
  Randomized noise shuffling for fast and consistent video editing with
  diffusion models,'' in \emph{CVPR}, 2024.

\bibitem{slicedit}
N.~Cohen, V.~Kulikov, M.~Kleiner, I.~Huberman-Spiegelglas, and T.~Michaeli,
  ``Slicedit: Zero-shot video editing with text-to-image diffusion models using
  spatio-temporal slices,'' in \emph{ICML}, 2024.

\bibitem{flowvid}
F.~Liang, B.~Wu, J.~Wang, L.~Yu, K.~Li, Y.~Zhao, I.~Misra, J.-B. Huang,
  P.~Zhang, P.~Vajda, and D.~Marculescu, ``{FlowVid}: Taming imperfect optical
  flows for consistent video-to-video synthesis,'' \emph{arXiv preprint
  arXiv:2312.17681}, 2023.

\bibitem{ccedit}
R.~Feng, W.~Weng, Y.~Wang, Y.~Yuan, J.~Bao, C.~Luo, Z.~Chen, and B.~Guo,
  ``{CCEdit}: Creative and controllable video editing via diffusion models,''
  in \emph{CVPR}, 2024.

\bibitem{motiondirector}
R.~Zhao, Y.~Gu, J.~Z. Wu, D.~J. Zhang, J.~Liu, W.~Wu, J.~Keppo, and M.~Z. Shou,
  ``{MotionDirector}: Motion customization of text-to-video diffusion models,''
  in \emph{ECCV}, 2024.

\bibitem{videoeditsurvey}
W.~Sun, R.-C. Tu, J.~Liao, and D.~Tao, ``Diffusion model-based video editing: A
  survey,'' \emph{arXiv preprint arXiv:2407.07111}, 2024.

\bibitem{svd}
A.~Blattmann, T.~Dockhorn, S.~Kulal, D.~Mendelevitch, M.~Kilian, D.~Lorenz,
  Y.~Levi, Z.~English, V.~Voleti, A.~Letts \emph{et~al.}, ``Stable video
  diffusion: Scaling latent video diffusion models to large datasets,''
  \emph{arXiv preprint arXiv:2311.15127}, 2023.

\bibitem{videocrafter}
H.~Chen, M.~Xia, Y.~He, Y.~Zhang, X.~Cun, S.~Yang, J.~Xing, Y.~Liu, Q.~Chen,
  X.~Wang, C.~Weng, and Y.~Shan, ``{VideoCrafter1}: Open diffusion models for
  high-quality video generation,'' \emph{arXiv preprint arXiv:2310.19512},
  2023.

\bibitem{dynamicrafter}
J.~Xing, M.~Xia, Y.~Zhang, H.~Chen, W.~Yu, H.~Liu, X.~Wang, T.-T. Wong, and
  Y.~Shan, ``{DynamiCrafter}: Animating open-domain images with video diffusion
  priors,'' in \emph{ECCV}, 2024.

\bibitem{lumiere}
O.~Bar-Tal, H.~Chefer, O.~Tov, C.~Herrmann, R.~Paiss, S.~Zada, A.~Ephrat,
  J.~Hur, G.~Liu, A.~Raj \emph{et~al.}, ``Lumiere: A space-time diffusion model
  for video generation,'' \emph{arXiv preprint arXiv:2401.12945}, 2024.

\bibitem{pisco}
\BIBentryALTinterwordspacing
X.~Gao, R.~Li, X.~Chen, Y.~Wu, S.~Feng, Q.~Yin, and Z.~Tu, ``{PISCO}: Precise
  video instance insertion with sparse control,'' \emph{arXiv preprint
  arXiv:2602.08277}, 2026. [Online]. Available:
  \url{https://arxiv.org/abs/2602.08277}
\BIBentrySTDinterwordspacing

\bibitem{sparkvsr}
\BIBentryALTinterwordspacing
J.~Yu, X.~Gao, P.~Verlani, A.~Gadde, Y.~Wang, B.~Adsumilli, and Z.~Tu,
  ``{SparkVSR}: Interactive video super-resolution via sparse keyframe
  propagation,'' in \emph{European Conference on Computer Vision}, 2026.
  [Online]. Available: \url{https://arxiv.org/abs/2603.16864}
\BIBentrySTDinterwordspacing

\bibitem{considgen}
\BIBentryALTinterwordspacing
M.~Wu, A.~Mishra, S.~Dey, S.~Xing, N.~Ravipati, H.~Wu, B.~Li, and Z.~Tu,
  ``{ConsID-Gen}: View-consistent and identity-preserving image-to-video
  generation,'' in \emph{Proceedings of the IEEE/CVF Conference on Computer
  Vision and Pattern Recognition}, 2026. [Online]. Available:
  \url{https://arxiv.org/abs/2602.10113}
\BIBentrySTDinterwordspacing

\bibitem{swaptext}
Q.~Yang, J.~Huang, and W.~Lin, ``{SwapText}: Image based texts transfer in
  scenes,'' in \emph{CVPR}, 2020.

\bibitem{mostel}
Y.~Qu, Q.~Tan, H.~Xie, J.~Xu, Y.~Wang, and Y.~Zhang, ``Exploring stroke-level
  modifications for scene text editing,'' in \emph{AAAI}, 2023.

\bibitem{glyphdraw}
J.~Ma, M.~Zhao, C.~Chen, R.~Wang, D.~Niu, H.~Lu, and X.~Lin, ``{GlyphDraw}:
  Seamlessly rendering text with intricate spatial structures in text-to-image
  generation,'' \emph{arXiv preprint arXiv:2303.17870}, 2023.

\bibitem{diffste}
J.~Ji, G.~Zhang, Z.~Wang, B.~Hou, Z.~Zhang, B.~L. Price, and S.~Chang,
  ``Improving diffusion models for scene text editing with dual encoders,''
  \emph{Transactions on Machine Learning Research (TMLR)}, 2024.

\bibitem{diffute}
H.~Chen, Z.~Xu, Z.~Gu, J.~Lan, X.~Zheng, Y.~Li, C.~Meng, H.~Zhu, and W.~Wang,
  ``{DiffUTE}: Universal text editing diffusion model,'' in \emph{NeurIPS},
  2023.

\bibitem{udifftext}
Y.~Zhao and Z.~Lian, ``{UDiffText}: A unified framework for high-quality text
  synthesis in arbitrary images via character-aware diffusion models,'' in
  \emph{ECCV}, 2024.

\bibitem{textdiffuser2}
J.~Chen, Y.~Huang, T.~Lv, L.~Cui, Q.~Chen, and F.~Wei, ``{TextDiffuser-2}:
  Unleashing the power of language models for text rendering,'' in \emph{ECCV},
  2024.

\bibitem{textflux}
Y.~Xie, J.~Zhang, P.~Chen, W.~Wang, L.~Gao, P.~Li, Q.~Qiao, and Z.~Lian,
  ``{TextFlux}: An ocr-free dit model for high-fidelity multilingual scene text
  synthesis,'' \emph{arXiv preprint arXiv:2505.17778}, 2025.

\bibitem{textmastero}
T.~Wang, X.~Qu, and T.~Liu, ``{TextMastero}: Mastering high-quality scene text
  editing in diverse languages and styles,'' \emph{arXiv preprint
  arXiv:2408.10623}, 2024.

\bibitem{strive}
{Vijay Kumar B G}, J.~Subramanian, V.~Chordia, E.~Bart, S.~Fang, K.~Guan, and
  R.~Bala, ``{STRIVE}: Scene text replacement in videos,'' in \emph{ICCV},
  2021.

\bibitem{vidtextpres}
Z.~Liu, K.~Valencia, and J.~Cui, ``Video text preservation with synthetic
  text-rich videos,'' \emph{arXiv preprint arXiv:2511.05573}, 2025.

\bibitem{legit}
M.~Mandal, N.~Birkbeck, B.~Adsumilli, and A.~C. Bovik, ``{LegiT}: Text
  legibility for user-generated media,'' in \emph{IEEE International Conference
  on Image Processing (ICIP)}, 2024.

\bibitem{vqasurvey}
\BIBentryALTinterwordspacing
Q.~Zheng, Y.~Fan, L.~Huang, T.~Zhu, J.~Liu, Z.~Hao, S.~Xing, C.-J. Chen,
  X.~Min, A.~C. Bovik, and Z.~Tu, ``Video quality assessment: A comprehensive
  survey,'' \emph{arXiv preprint arXiv:2412.04508}, 2024. [Online]. Available:
  \url{https://arxiv.org/abs/2412.04508}
\BIBentrySTDinterwordspacing

\bibitem{cover}
\BIBentryALTinterwordspacing
C.~He, Q.~Zheng, R.~Zhu, X.~Zeng, Y.~Fan, and Z.~Tu, ``{COVER}: A comprehensive
  video quality evaluator,'' in \emph{Proceedings of the IEEE/CVF Conference on
  Computer Vision and Pattern Recognition Workshops}, 2024, pp. 5799--5809.
  [Online]. Available:
  \url{https://openaccess.thecvf.com/content/CVPR2024W/AI4Streaming/html/He_COVER_A_Comprehensive_Video_Quality_Evaluator_CVPRW_2024_paper.html}
\BIBentrySTDinterwordspacing

\bibitem{physiq}
S.~Motamed, L.~Culp, K.~Swersky, P.~Jaini, and R.~Geirhos, ``Do generative
  video models understand physical principles?'' \emph{arXiv preprint
  arXiv:2501.09038}, 2025.

\bibitem{openve}
H.~He, J.~Wang, J.~Zhang, Z.~Xue, X.~Bu, Q.~Yang, S.~Wen, and L.~Xie,
  ``{OpenVE-3M}: A large-scale high-quality dataset for instruction-guided
  video editing,'' \emph{arXiv preprint arXiv:2512.07826}, 2025.

\bibitem{tdvea}
J.~Wang, J.~Wang, H.~Duan, G.~Zhai, and X.~Min, ``{TDVE-Assessor}: Benchmarking
  and evaluating the quality of text-driven video editing with {LMM}s,''
  \emph{arXiv preprint arXiv:2505.19535}, 2025.

\bibitem{sam3}
N.~Carion, L.~Gustafson, Y.-T. Hu \emph{et~al.}, ``{SAM 3}: Segment anything
  with concepts,'' \emph{arXiv preprint arXiv:2511.16719}, 2025.

\bibitem{qwen3vl}
{Qwen Team, Alibaba Cloud}, ``{Qwen3-VL}: Vision-language foundation model,''
  \url{https://huggingface.co/Qwen/Qwen3-VL-32B-Instruct}, 2025.

\bibitem{rose}
C.~Miao, Y.~Feng, J.~Zeng, Z.~Gao, H.~Liu, Y.~Yan, D.~Qi, X.~Chen, B.~Wang, and
  H.~Zhao, ``{ROSE}: Remove objects with side effects in videos,'' \emph{arXiv
  preprint arXiv:2508.18633}, 2025.

\bibitem{nanobananapro}
{Google DeepMind}, ``Gemini 3 pro image (``nano banana pro''),''
  \url{https://deepmind.google/models/gemini-image/pro/}, 2025.

\bibitem{ppocrv5}
C.~Cui, T.~Sun, M.~Lin, T.~Gao, Y.~Zhang, J.~Liu, X.~Wang, Z.~Zhang, C.~Zhou,
  H.~Liu, Y.~Zhang, W.~Lv, K.~Huang, Y.~Zhang, J.~Zhang, J.~Zhang, Y.~Liu,
  D.~Yu, and Y.~Ma, ``{PaddleOCR 3.0} technical report,'' \emph{arXiv preprint
  arXiv:2507.05595}, 2025.

\bibitem{raft}
Z.~Teed and J.~Deng, ``{RAFT}: Recurrent all-pairs field transforms for optical
  flow,'' in \emph{European Conference on Computer Vision (ECCV)}, 2020, pp.
  402--419.

\bibitem{musiq}
J.~Ke, Q.~Wang, Y.~Wang, P.~Milanfar, and F.~Yang, ``{MUSIQ}: Multi-scale image
  quality transformer,'' in \emph{ICCV}, 2021.

\bibitem{ssim}
Z.~Wang, A.~C. Bovik, H.~R. Sheikh, and E.~P. Simoncelli, ``Image quality
  assessment: From error visibility to structural similarity,'' \emph{IEEE
  Transactions on Image Processing}, vol.~13, no.~4, pp. 600--612, 2004.

\bibitem{lpips}
R.~Zhang, P.~Isola, A.~A. Efros, E.~Shechtman, and O.~Wang, ``The unreasonable
  effectiveness of deep features as a perceptual metric,'' in \emph{CVPR},
  2018.

\bibitem{dreamsim}
S.~Fu, N.~Tamir, S.~Sundaram, L.~Chai, R.~Zhang, T.~Dekel, and P.~Isola,
  ``{DreamSim}: Learning new dimensions of human visual similarity using
  synthetic data,'' in \emph{Advances in Neural Information Processing Systems
  (NeurIPS)}, 2023.

\bibitem{datasheets}
T.~Gebru, J.~Morgenstern, B.~Vecchione, J.~W. Vaughan, H.~Wallach,
  H.~Daum{\'e}~III, and K.~Crawford, ``Datasheets for datasets,''
  \emph{Communications of the ACM}, 2021.

\bibitem{croissant}
M.~Akhtar, O.~Benjelloun, C.~Conforti \emph{et~al.}, ``{Croissant}: A metadata
  format for ml-ready datasets,'' in \emph{DEEM Workshop @ SIGMOD}, 2024.

\bibitem{depthanything3}
H.~Lin, S.~Chen, J.~Liew, D.~Y. Chen, Z.~Li, G.~Shi, J.~Feng, and B.~Kang,
  ``Depth anything 3: Recovering the visual space from any views,'' \emph{arXiv
  preprint arXiv:2511.10647}, 2025.

\bibitem{easyocr}
{Jaided AI}, ``{EasyOCR}: Ready-to-use {OCR} with 80+ supported languages,''
  \url{https://github.com/JaidedAI/EasyOCR}, 2020.

\bibitem{cotracker3}
N.~Karaev, Y.~Makarov, J.~Wang, N.~Neverova, A.~Vedaldi, and C.~Rupprecht,
  ``{CoTracker3}: Simpler and better point tracking by pseudo-labelling real
  videos,'' in \emph{ICCV}, 2025.

\bibitem{pulseofmotion}
\BIBentryALTinterwordspacing
X.~Gao, M.~Wu, S.~Yang, J.~Yu, P.~Taghavi, F.~Lin, and Z.~Tu, ``The pulse of
  motion: Measuring physical frame rate from visual dynamics,'' \emph{arXiv
  preprint arXiv:2603.14375}, 2026. [Online]. Available:
  \url{https://arxiv.org/abs/2603.14375}
\BIBentrySTDinterwordspacing

\bibitem{dinov2}
\BIBentryALTinterwordspacing
M.~Oquab, T.~Darcet, T.~Moutakanni, H.~Vo, M.~Szafraniec, V.~Khalidov,
  P.~Fernandez, D.~Haziza, F.~Massa, A.~El-Nouby, M.~Assran, N.~Ballas,
  W.~Galuba, R.~Howes, P.-Y. Huang, S.-W. Li, I.~Misra, M.~Rabbat, V.~Sharma,
  G.~Synnaeve, H.~Xu, H.~Jegou, J.~Mairal, P.~Labatut, A.~Joulin, and
  P.~Bojanowski, ``{DINOv2}: Learning robust visual features without
  supervision,'' \emph{arXiv preprint arXiv:2304.07193}, 2023. [Online].
  Available: \url{https://arxiv.org/abs/2304.07193}
\BIBentrySTDinterwordspacing

\bibitem{retinaface}
\BIBentryALTinterwordspacing
J.~Deng, J.~Guo, E.~Ververas, I.~Kotsia, and S.~Zafeiriou, ``{RetinaFace}:
  Single-shot multi-level face localisation in the wild,'' in \emph{Proceedings
  of the IEEE/CVF Conference on Computer Vision and Pattern Recognition}, 2020.
  [Online]. Available:
  \url{https://openaccess.thecvf.com/content_CVPR_2020/html/Deng_RetinaFace_Single-Shot_Multi-Level_Face_Localisation_in_the_Wild_CVPR_2020_paper.html}
\BIBentrySTDinterwordspacing

\bibitem{arcface}
\BIBentryALTinterwordspacing
J.~Deng, J.~Guo, N.~Xue, and S.~Zafeiriou, ``{ArcFace}: Additive angular margin
  loss for deep face recognition,'' in \emph{Proceedings of the IEEE/CVF
  Conference on Computer Vision and Pattern Recognition}, 2019. [Online].
  Available:
  \url{https://openaccess.thecvf.com/content_CVPR_2019/html/Deng_ArcFace_Additive_Angular_Margin_Loss_for_Deep_Face_Recognition_CVPR_2019_paper.html}
\BIBentrySTDinterwordspacing

\end{thebibliography}
